\documentclass[10pt,twocolumn,letterpaper]{article}

\usepackage[pagenumbers]{cvpr} 

\usepackage{cuted}
\usepackage{caption}
\usepackage{multirow}
\usepackage{array}
\usepackage{colortbl}
\usepackage{adjustbox}

\definecolor{cvprblue}{rgb}{0.21,0.49,0.74}
\usepackage[pagebackref,breaklinks,colorlinks,allcolors=cvprblue]{hyperref}

\def\confName{CVPR}
\def\confYear{2027}

\title{A Failure-Driven Data Flywheel for Post-Training Vision-Language-Action Models}

\author{
Yuhao Luo$^{1,*}$ \quad Zihao Sheng$^{1,*}$ \quad Zilin Huang$^{1}$ \quad Zhengyang Wan$^{1}$\\
Boyue Wang$^{1}$ \quad Ruolin Yang$^{1}$ \quad Shuai Wang$^{2}$ \quad Sikai Chen$^{1,\dagger}$\\[0.5em]
$^{1}$University of Wisconsin--Madison \qquad $^{2}$Southeast University\\[0.3em]
{\small $^{*}$Equal contribution \qquad $^{\dagger}$Corresponding author: \texttt{sikai.chen@wisc.edu}}
}

\begin{document}
\maketitle

\begin{strip}
    \centering
    \includegraphics[width=\linewidth]{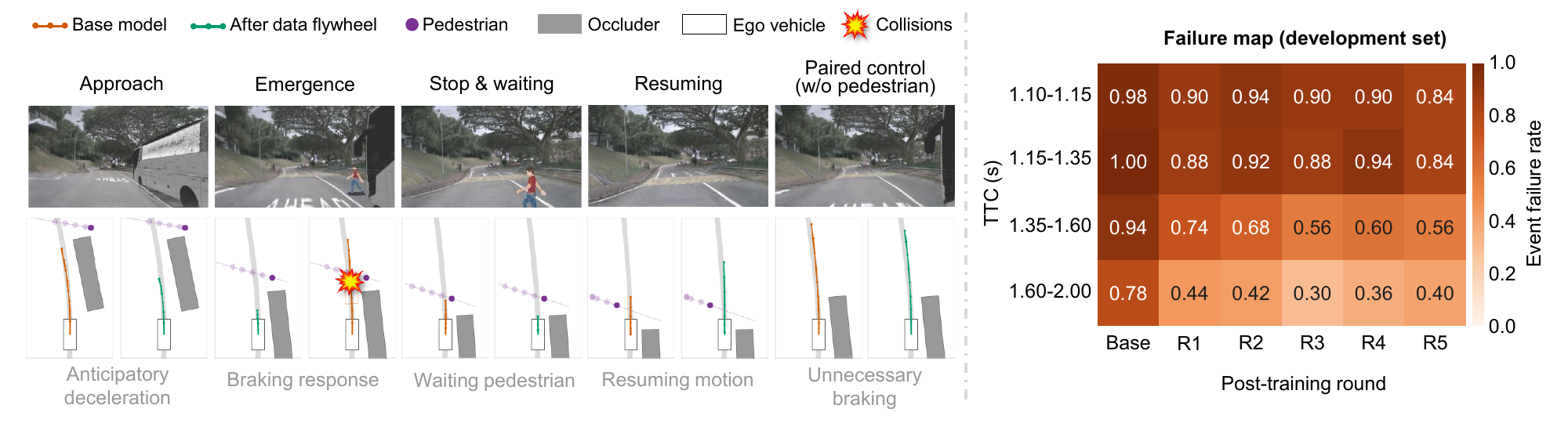}
    \captionsetup{hypcap=false}
    \captionof{figure}{\textbf{Failure-driven data flywheel on ReCogDrive.}
    \textbf{(Left)} Open-loop \(5\,\mathrm{s}\) plans before (orange) and after
    (green) post-training for an occluded-pedestrian scenario and its paired control.
    \textbf{(Right)} Failure map of ReCogDrive on the development set.
The map reports the event failure rate within each time to collision (TTC) bin, where an event fails if a predicted collision with the pedestrian occurs at any critical-phase query.}
    \label{fig:teaser}
\end{strip}

\begin{abstract}
Pretrained vision-language-action (VLA) models rarely satisfy all behavioral requirements of a new platform, and teams without a large-scale deployed fleet lack the data required to address the resulting failures.
To address this problem, we propose a failure-driven data flywheel for post-training such models.
The framework constructs target scenarios absent from the available data by inserting task-relevant entities into reconstructed real-world scenes, and evaluates the model in these scenarios against the individual behavioral requirements of the target task.
The resulting failure rates form a failure map across task conditions and difficulty levels.
The map is updated through re-evaluation after each round and assigns larger data generation quotas to conditions under which the model continues to fail. We instantiate the framework in autonomous driving for occluded-pedestrian emergence and yielding at stop-sign-controlled intersections, and evaluate it on two VLA models and one non-VLA planner in both simulation and real-vehicle experiments.
For occluded-pedestrian emergence, five rounds of post-training improve the critical-phase collision-free rate on the sealed test set from 0.261 to 0.579 for AutoVLA and from 0.390 to 0.765 for ReCogDrive, while the unnecessary braking rate in the paired control scenarios never exceeds 0.013. When deployed on a full-scale vehicle without additional real-world training, the post-trained ReCogDrive model achieves 4 of 5 safe passages in the target scenario, compared with 1 of 5 for the base model.
Fine-grained evaluation further reveals a behavioral pattern obscured by the collision-free rate alone: both VLA models acquire anticipatory deceleration before reaching the occluding vehicle, whereas their braking response after pedestrian emergence remains limited.
For yielding at stop-sign-controlled intersections, the same framework increases the stop-before-line rate from 0.100 to 0.897 for AutoVLA and from 0.000 to 0.667 for ReCogDrive.
Under sequential training initialized from the pedestrian-capability models, the stop-before-line rate reaches 0.599 and 0.358, respectively, while the pedestrian collision-free rate decreases by at most 0.032.
\end{abstract}
\section{Introduction}
\label{sec:intro}
Vision-language-action (VLA) models integrate multimodal understanding with action generation~\cite{driess2023palm, brohan2023rt, kim2024openvla} and have been applied to robotic manipulation~\cite{ma2026survey, zhao2025cot, zhao2025vlas} and autonomous driving~\cite{yuan2026autodrive, liu2026driveworld, liu2026uni}. 
With the public release of pretrained VLA models, a growing number of teams adapt existing models to their own platforms and tasks without training them from scratch. 
Even after domain-specific training, however, these models may still fail to meet the requirements of specific tasks. 

Addressing these residual failures poses two challenges: identifying which capabilities a model lacks and acquiring targeted training data. 
For evaluation, driving capabilities have been assessed in synthetic environments and through closed-loop safety testing in reconstructed real-world scenes~\cite{jia2024bench2drive,ljungbergh2024neuroncap,zhou2025hugsim}, yet results reported by scenario category or through aggregate metrics may still obscure which individual behavioral requirements remain unmet~\cite{moayeri2025unearthing, ye2024flask, zhang2025vla, yang2026beyond}. 
For training, iterative learning has been explored using generated reasoning traces and interaction experience generated by world models~\cite{zelikman2022star,yang2026rise}, and several studies further use behavioral analysis or failure signals to guide the acquisition and selection of training data for driving models~\cite{sheng2026curricuvlm,li2026world,gao2026learning,dou2026fire}. 
These studies identify failures from existing data or from the behavior of the policy on such data. 
When the behaviors required by a new platform have no corresponding events in the available data~\cite{xu2025can, sagar2026robomd, pumacay2024colosseum, li2025autobencher}, such failures remain unexposed, particularly for teams without a large-scale deployed fleet. 
Moreover, failures may shift as post-training proceeds~\cite{sheng2026curricuvlm, jia2026blind, yu2025self}, motivating re-evaluation and corresponding adjustments to data generation across training rounds. 

To address these challenges, we propose a failure-driven data flywheel for post-training VLA models. 
The framework evaluates individual behavioral requirements, constructs target scenarios under controlled conditions, and uses expert demonstrations as supervision. 
On a development set, event failure rates across task conditions and difficulty levels form a failure map, which the framework uses to assign larger data generation quotas to conditions with higher failure rates. 
After each round of post-training, the model is re-evaluated and the failure map is updated for the subsequent round. 
Generated data are mixed with original-task data to support the retention of existing capabilities, while final performance is evaluated on an independent sealed test set excluded from all iterative decisions. 
We instantiate the framework in autonomous driving with two target capabilities: occluded-pedestrian emergence and yielding at stop-sign-controlled intersections. 
To implement controllable data generation, we reconstruct real-world scenes from nuScenes logs~\cite{caesar2020nuscenes} using 3D Gaussian Splatting~\cite{10.1145/3592433, zhou2024drivinggaussian, hess2025splatad, chen2025omnire} and insert 3D assets~\cite{ljungbergh2026r3d2, yang2025genassets} representing traffic participants and traffic signs. 
By controlling asset placement and participant motion timing, we construct scenarios absent from the source logs with controllable interaction conditions and difficulty levels. 
A rule-based expert generates feasible trajectories that satisfy the target behavioral requirements, and observations rendered along these trajectories are paired with the corresponding future trajectory segments as supervision targets. 
Paired control scenarios identify unnecessary braking or stopping. 
We evaluate the framework on three publicly available driving models: two VLA models, AutoVLA~\cite{zhou2026autovla} and ReCogDrive~\cite{ICLR2026_ff7bf601}, and a non-VLA planner, LAW~\cite{li2025enhancing}, which serves to examine whether the framework depends on the VLA architecture. 

On the sealed test set for occluded-pedestrian emergence, five rounds of post-training improve the critical-phase collision-free rate from 0.261 to 0.579 for AutoVLA and from 0.390 to 0.765 for ReCogDrive, while the unnecessary braking rate in the paired control scenarios never exceeds 0.013. 
Fine-grained evaluation further reveals behavioral patterns obscured by the collision-free rate alone. 
Both VLA models acquire anticipatory deceleration before reaching the occluding vehicle, whereas their braking response after pedestrian emergence remains limited (Figure~\ref{fig:teaser}, left). 
Moreover, removing the paired control scenario samples increases the collision-free rate to 0.944, yet this apparent improvement arises from indiscriminate braking. 
The improvement is also observed on a full-scale vehicle, where the number of safe passages of ReCogDrive increases from 1 of 5 to 4 of 5 without additional real-world training.

Our main contributions are as follows: 
\begin{itemize}
    \item We propose a failure-driven data flywheel for post-training VLA models. After each round, bin-wise event failure rates update the failure map and determine the data generation quotas for the next round. 
    \item We introduce a fine-grained evaluation that decomposes each target scenario into stage-specific behavioral requirements. Together with paired target and control scenarios, these requirements distinguish appropriate event responses from overly conservative strategies. 
    \item Experiments on three driving models and two target capabilities demonstrate improvements in both open- and closed-loop settings, and the improvement is also observed in real-vehicle experiments on ReCogDrive. Under matched training budgets, the proposed flywheel reduces the open-loop critical-phase collision rate by 10.1\% relative to the strongest baseline, and sequential experiments demonstrate the acquisition of a new capability while largely retaining the previously acquired capability. 
\end{itemize}

\section{Related Work}
\label{sec:related}

\noindent\textbf{Fine-grained capability evaluation.}
In autonomous driving, Bench2Drive~\cite{jia2024bench2drive} organizes closed-loop tests in CARLA around multiple driving abilities, advancing evaluation beyond overall driving scores toward capability-specific assessment. 
Neural rendering further enables policy testing in reconstructed real-world scenes. 
NeuroNCAP~\cite{ljungbergh2024neuroncap} uses a neural radiance field (NeRF) to construct editable safety-critical scenarios, while HUGSIM~\cite{zhou2025hugsim} employs 3D Gaussian Splatting for photorealistic closed-loop simulation and evaluates driving policies through challenging traffic interactions. 
We further examine whether individual behavioral requirements within a target scenario are satisfied and use paired controls to distinguish appropriate event responses from overly conservative strategies.

\noindent\textbf{Synthetic training data and feedback-driven post-training.}
Simulation and generative models provide training data for policy learning beyond real-world data collection. 
SimScale~\cite{tian2026simscale} generates new driving states by perturbing the ego trajectory in reconstructed real-world scenes and provides supervision using pseudo-expert trajectories.
World Engine~\cite{li2026world} identifies failure-prone scenarios in real-world logs and combines scene reconstruction with traffic behavior generation for post-training. 
Both methods derive their scenarios from existing driving logs, whereas our framework inserts new traffic participants into reconstructed real-world scenes to instantiate prespecified target scenarios that are absent from the source logs, with controllable interaction conditions and difficulty.
CurricuVLM~\cite{sheng2026curricuvlm} and CorrectAD~\cite{ma2026correctad} use behavioral and failure analyses, respectively, to guide data generation.
AutoScale~\cite{ruan2026closed} selects samples from a pre-generated simulation pool and adjusts the training data mixture based on evaluation feedback. 
TakeVLA~\cite{gao2026learning} and FIRE-VLA~\cite{dou2026fire} iteratively update driving VLA models using expert takeover events and low-reward trajectories, respectively. 
In these studies, the failure signals are obtained from existing driving data or from the behavior of the policy on such data. 
In our framework, they are obtained from fine-grained behavioral evaluation of the constructed scenarios after each training round and update the failure maps that guide subsequent data generation.

\section{Method}
\label{sec:method}

\begin{figure*}[t]
    \centering
    \includegraphics[width=0.98\linewidth]{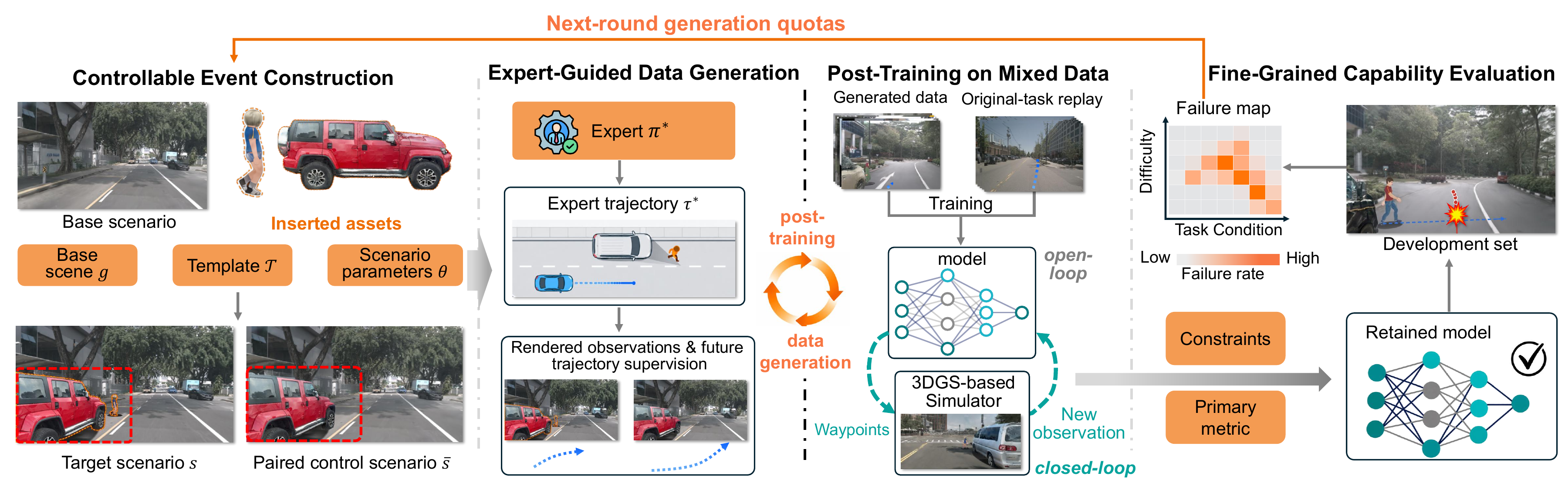}
    \caption{%
Overview of the proposed failure-driven data flywheel. 
After each round of post-training, bin-wise event failure rates determine the data generation quotas for the next round. 
}
    \label{fig:framework}
\end{figure*}

Figure~\ref{fig:framework} provides an overview of the proposed failure-driven data flywheel. 
Given a pretrained model and a prespecified target capability, the framework initializes the first round by generating target scenarios with uniform quotas across task-condition and difficulty-level bins. 
At each round, the newly generated scenarios are incorporated into the post-training mixture, and post-training is run to convergence, producing a sequence of candidate checkpoints across epochs. 
These checkpoints are evaluated on a fixed development set to determine the checkpoint retained for that round (Section~\ref{sec:flywheel-evaluation}). 
Its bin-wise event failure rates are then used to update the failure map, which governs data generation quotas for the next round. 
This iterative process forms a feedback loop among capability evaluation, targeted data generation, and post-training. 

We use occluded-pedestrian emergence as a running example and describe the stop-sign yielding instantiation in Appendix~B. 

\subsection{Problem Setup}
\label{sec:flywheel-setup}

We specify a target capability as \(c=(\mathcal{T},\Theta,\pi^\star,\mathcal{F})\), where \(\mathcal{T}\) is a scenario template, \(\Theta\) is the space of scenario parameters, \(\pi^\star\) is an expert that provides feasible demonstrations, and \(\mathcal{F}=(F_1,\ldots,F_k)\) is a tuple of capability metrics with \(F_1\) as the primary metric. 
A base scene \(g\) provides the environment in which an event is instantiated. 
The template $\mathcal{T}$ uses a feasible configuration \(\theta \in \Theta_{\mathrm{feas}}(g)\subseteq\Theta\) to construct a target scenario \(s=\mathcal{T}(g,\theta)\) and a paired control scenario \(\bar{s}\). 
The same specification can describe capabilities outside driving. 
In robotic grasping, for instance, \(\mathcal{T}\) could be a tabletop scene template, \(\Theta\) the space of object placements and the timing of their appearance, \(\pi^\star\) teleoperation, and \(\mathcal{F}\) the grasp success rate and the rate of unnecessary stops. 

\subsection{Controllable Event Construction}
\label{sec:flywheel-event-construction}

Given a base scene \(g\) and a feasible scenario parameter configuration \(\theta \in \Theta_{\mathrm{feas}}(g)\), the scenario template \(\mathcal{T}\) constructs a target scenario \(s=\mathcal{T}(g,\theta)\). 
The parameters \(\theta\) govern the placement of task-relevant entities and the motion of dynamic actors, including the timing of their interactions. 
In our driving experiments, we reconstruct base scenes $g$ from real-world driving logs using OmniRe~\cite{chen2025omnire}, which is based on 3D Gaussian Splatting~\cite{zhou2024drivinggaussian, khan2025autosplat, sun2025splatflow}. 
For occluded-pedestrian emergence, we insert an occluding vehicle and script a pedestrian to emerge from behind it into the ego vehicle's path. 
Appendix~B details the scenario template $\mathcal{T}$, the scenario parameters $\theta$, and the constraints that define $\Theta_{\mathrm{feas}}(g)$ for this task. 

In the pedestrian task, we parameterize event difficulty by the target time to collision (TTC), \(d(\theta)\), at first pedestrian visibility. 
Given a nominal ego approach at speed $v_{\mathrm{safe}}$, we solve backward from the desired TTC to obtain the nominal ego position at emergence $u_{\mathrm{rev}}$ and the corresponding pedestrian departure time $t_{\mathrm{depart}}$: 
\begin{equation}
\begin{aligned}
u_{\mathrm{rev}} &= u_{\mathrm{conf}} - \frac{L}{2}
    - v_{\mathrm{safe}}d(\theta),\\
t_{\mathrm{depart}} &= t_{\mathrm{rev}}^{\mathrm{nom}}
    - \frac{\ell_{\mathrm{vis}}}{v_p}.
\end{aligned}
\label{eq:flywheel-event-timing}
\end{equation}
Here, \(u_{\mathrm{conf}}\) is the conflict-point position along the ego path, \(L\) is the ego-vehicle length, and \(t_{\mathrm{rev}}^{\mathrm{nom}}\) is the time at which the nominal ego motion reaches \(u_{\mathrm{rev}}\). 
The pedestrian travels a distance \(\ell_{\mathrm{vis}}\) at speed \(v_p\) before becoming visible. 
The nominal ego motion establishes the timing of the scripted pedestrian scenario. 
The expert driving trajectory is generated after the scenario has been specified. 

The template \(\mathcal{T}\) also constructs a paired control scenario \(\bar{s}\) by removing the factor that triggers the target response while preserving the remaining configuration of the target scenario \(s\). 
For occluded-pedestrian emergence, \(\bar{s}\) retains the same base scene and occluding-vehicle placement as \(s\), but omits the pedestrian. 

\subsection{Expert-Guided Data Generation}
\label{sec:flywheel-expert-data}

Any source of feasible demonstrations, such as a motion planner, teleoperation, or human intervention, can serve as the expert \(\pi^\star\). 
In our driving experiments, we use a rule-based expert. 
Given the constructed pair, the expert generates separate demonstrations \(\tau^\star=\pi^\star(s)\) and \(\bar{\tau}^\star=\pi^\star(\bar{s})\). 
In the pedestrian task, the expert follows the recorded route and decelerates before reaching the occluder in both scenarios. 
In the target scenario $s$, it brakes after pedestrian emergence and a reaction delay, stops before the conflict, and resumes motion once the pedestrian clears the path. 
In the paired control scenario $\bar{s}$, it proceeds after the shared anticipatory deceleration. 
Appendix~B details the expert speed profiles and parameters. 

We render camera observations along each expert trajectory and use the corresponding future trajectory segment as the supervision target at each sampled time. 
Thus, the observation and future waypoints are aligned at the same state and timestamp. 

\subsection{Post-Training on Mixed Data}
\label{sec:flywheel-post-training}

Each round starts from the checkpoint retained in the previous round, with the pretrained model \(\pi_0\) used to initialize the first round. 
The generated-data pool in round $r$, denoted by $\mathcal{D}_r^{\mathrm{gen}}$, comprises supervision samples derived from the current round's generated target scenarios $s$ and paired control scenarios $\bar s$, scenarios rendered from the same base scenes without any inserted event, and, for $r>1$, a limited replay subset drawn from previous rounds. 
We mix this pool with original-task data, which in our driving experiments consist of samples from real-world driving logs. 
The training distribution in round \(r\) is 
\begin{equation}
p_r^{\mathrm{train}}
  =(1-\rho)p_r^{\mathrm{gen}}+\rho p^{\mathrm{orig}},
\label{eq:flywheel-training-mixture}
\end{equation}
where \(p_r^{\mathrm{gen}}\) is the sampling distribution over \(\mathcal{D}_r^{\mathrm{gen}}\), \(p^{\mathrm{orig}}\) is the sampling distribution over the original-task data, and \(\rho\) is the original-task replay proportion. 
Each round trains to convergence over multiple epochs. 
Checkpoints saved during optimization form the candidate set for that round. 

\subsection{Fine-Grained Capability Evaluation}
\label{sec:flywheel-evaluation}

\noindent\textbf{Fine-grained behavioral evaluation.}
For each target capability, we define query states and metrics for evaluating the required behaviors in the target scenario \(s\) and its paired control \(\bar{s}\). 
At each query time \(t\), the model predicts a future trajectory \(\hat{\tau}_t=\pi(o_t)\) from observation \(o_t\). 
The metrics assess these predictions against the behavioral requirements, using expert trajectories and scripted actor motion as references. 
The primary metric \(F_1\) ranks candidate checkpoints, while the other metrics characterize individual behaviors. 
Selected metrics serve as selection constraints. 

For occluded-pedestrian emergence, we assess collision avoidance throughout the critical phase, which spans the approach to the occluded region, pedestrian emergence, and the ensuing braking response. 
In the target scenario \(s\), we examine whether the ego vehicle decelerates before the pedestrian becomes visible and applies additional braking after emergence. 
We also assess whether it remains stopped while the conflict persists and resumes motion after the pedestrian clears its path. 
In the paired control scenario \(\bar{s}\), we assess whether the ego vehicle makes forward progress without unnecessary braking when no pedestrian emerges. 

\noindent\textbf{Checkpoint selection.}
At each round, we evaluate the candidate checkpoints on a fixed development set \(\mathcal{S}^{\mathrm{dev}}\) and retain the one with the highest primary metric \(F_1\) among those that satisfy the behavioral constraints and a predefined bound on the original-task validation error. 
Section~4.1 lists the thresholds, and Appendix~B details the complete selection and acceptance rule. 

\noindent\textbf{Failure map and generation quotas.}
Let \(\mathcal{B}\) denote the set of task-condition and difficulty-level bins. 
A grouping function \(\phi(g,\theta)\) assigns each target scenario in \(\mathcal{S}^{\mathrm{dev}}\) to a bin \(b\in\mathcal{B}\). 
For occluded-pedestrian emergence, \(\mathcal{B}\) comprises four bins defined solely by the target TTC \(d(\theta)\), and an event fails if a predicted ego--pedestrian collision occurs at any critical-phase query. 
After round \(r\), we compute the failure rate \(m_r(b)\) of the retained checkpoint within each bin. 
These rates form the failure map (Figure~\ref{fig:teaser}, right) and determine the proportion \(q_{r+1}(b)\) of the next round's generation budget allocated to bin \(b\). 
\begin{equation}
q_{r+1}(b)=
\frac{\max\{m_r(b),\epsilon_q\}}
     {\sum_{b'\in\mathcal{B}}\max\{m_r(b'),\epsilon_q\}},
\qquad r\geq1,
\label{eq:flywheel-generation-quotas}
\end{equation}
where \(\epsilon_q>0\) gives every bin a positive allocation. 

\section{Experiments}

\label{sec:experiments}
\subsection{Experimental Setup}
\label{sec:experimental-setup}

\paragraph{Datasets.}
For occluded-pedestrian emergence, we use 140 base scenes from nuScenes~\cite{caesar2020nuscenes}, reconstructed using OmniRe~\cite{chen2025omnire}, a 3D Gaussian Splatting framework for dynamic urban scene reconstruction. We partition the scenes by source log into three disjoint
subsets, with 100 training scenes from the official training split and 20 development
and 20 sealed test scenes from the official validation split. The development and
sealed test sets each contain 200 fixed target scenarios, with 50 scenarios in each of the
four TTC bins \([1.10,1.15)\), \([1.15,1.35)\), \([1.35,1.60)\), and \([1.60,2.00]\)~s.
TTC is sampled uniformly within each bin. Each target scenario is matched to a
control from the same base scene and timestamp, yielding 200 pairs per set.

Each target scenario inserts one pedestrian and one occluding vehicle. Adult pedestrians
use THuman2.1~\cite{tao2021function4d} scans animated with frame-wise SMPL-X~\cite{pavlakos2019expressive} poses recovered from pedestrians in
the corresponding nuScenes log. Child pedestrians are eight self-built rigged avatars, created with MakeHuman~\cite{bastioni2008ideas} and Microsoft Rocketbox~\cite{gonzalez2020rocketbox}. Occluding vehicles are drawn primarily from Objaverse~\cite{deitke2023objaverse} and supplemented with scanned vehicles from 3DRealCar~\cite{du20253drealcar}. Appendix~B gives the asset screening criteria and usage proportions. We construct 500 new target scenarios in the first round and 400 in each subsequent round, all from the training scenes and each with a paired control. Observations are rendered
at 10~Hz from three forward-facing cameras at \(800\times450\) pixels.

\noindent\textbf{Models.}
AutoVLA~\cite{zhou2026autovla} uses Qwen2.5-VL-3B~\cite{qwen2.5-VL, Qwen2VL, Qwen-VL} and takes three front-facing camera views, ego state, and a
navigation instruction as input. We initialize it from the official weights and follow
the official supervised fine-tuning recipe, freezing the vision encoder and updating
the language model. ReCogDrive~\cite{ICLR2026_ff7bf601} combines a frozen InternVL3-2B~\cite{chen2024expanding, wang2024mpo} with a diffusion planner
and takes one front-facing camera view, a four-step ego pose history, and a navigation
command. We update only the planner. LAW~\cite{li2025enhancing} is a non-VLA latent world model planner that
uses camera inputs without ego state. We adapt its official six-camera configuration to
the three available camera views. The initial ReCogDrive planner and LAW models are
trained on the nuScenes training split. Appendix~B details model-specific observation
histories, prediction horizons, and waypoint counts.

\noindent\textbf{Evaluation protocol and metrics.}
\textbf{Open loop.} Rather than measuring displacement errors, the open-loop capability metrics assess whether predicted ego trajectories at fixed query states satisfy the behavioral requirements described in Section~\ref{sec:flywheel-evaluation}. The \emph{primary} metric is the
critical-phase collision-free rate (CFR). \emph{Anticipatory Deceleration} (AD)
measures slowing on approach to the occluded region before the pedestrian
becomes visible. \emph{Braking Response} (BR) measures additional braking after
pedestrian emergence using matched target and control predictions.
\emph{Remaining Stopped} (RS) measures whether the vehicle holds its stop while
the conflict persists, and \emph{Resuming Motion} (RM) measures whether it resumes
driving after the conflict clears. In the paired control, \emph{Unnecessary Braking Rate} (UBR) measures excessive braking, and \emph{Control-Scenario Progress} (CSP)
measures forward progress. We compute the original-task validation error
\(L_{\mathrm{val}}\) on real frames from the full nuScenes validation split.

On the development set, checkpoint selection requires a UBR of at most \(0.10\) and an RS score of at least \(0.70\). We also require
\(L_{\mathrm{val}}(\pi)\leq 1.10L_{\mathrm{val}}(\pi_0)\).
We report open-loop capability scores on the sealed test pairs. Appendix B provides the full metric definitions and details the open-loop post-training procedure.

\textbf{Closed loop.} We evaluate episode-level safety and progress
over 163 sealed test pairs. The policy replans every \(0.5\)~s,
executes its first predicted waypoint, and receives a new observation
rendered at the resulting ego state. In the target scenario,
\emph{Pedestrian Collision} (PC) measures the fraction of episodes with a
pedestrian collision, \emph{Safe Passage} (SP) measures the fraction in which
the ego vehicle clears the conflict point without a terminal event, and
\emph{Progress} (P) measures normalized forward travel. In the paired
control, \emph{No-Event Rate} (NER) measures the fraction of episodes without
a terminal event, and \emph{Progress} (P) measures normalized forward travel.

For checkpoint selection, we use one minus the pedestrian collision rate
as the primary metric. Candidate checkpoints must achieve at least \(90\%\)
of the paired-control progress of the checkpoint used to initialize
closed-loop training and satisfy the same \(L_{\mathrm{val}}\) bound.
Appendix B also details closed-loop post-training.

\subsection{Main Results}

\noindent\textbf{Open-loop performance.} Table~\ref{tab:main_results} presents the open-loop results for three driving models from their base models through Round~5. These target scenarios are absent from the nuScenes logs, so the low base-model scores are expected and carry no implication for performance on the original benchmarks. For AutoVLA, the primary metric increases from 0.261 to 0.579, anticipatory deceleration improves from 0.100 to 0.575, and control-scenario progress rises from 0.751 to 0.914. Meanwhile, the validation error on the original tasks decreases by 13.2\%. For ReCogDrive, the primary metric increases from 0.390 to 0.765, anticipatory deceleration improves from 0.297 to 0.631, and remaining stopped reaches 0.976, whereas control-scenario progress decreases from 0.884 to 0.733 with the unnecessary braking rate remaining at 0.001. Its validation error on the original tasks increases by only 3.1\%. For both models, the improvement arises mainly from anticipatory deceleration before reaching the occluding vehicle, consistent with the expert, which decelerates before the occluder in both the target and paired control scenarios (Section~\ref{sec:flywheel-expert-data}). Braking response after pedestrian emergence also improves (0.090 to 0.115 for AutoVLA, 0.045 to 0.185 for ReCogDrive) yet remains limited, a pattern obscured by the collision-free rate alone. For both models, the data flywheel improves the target capability through multi-round post-training while satisfying the constraints and the requirements for retaining existing capabilities.

LAW follows a different pattern. The base model of our three-camera variant of LAW attains a collision-free rate of 0.967, accompanied by an unnecessary braking rate of 1.000 and control-scenario progress of only 0.113. These results indicate that its high collision-free rate arises mainly from policy stalling. After five rounds of post-training, LAW attains a collision-free rate of 0.791, reduces the unnecessary braking rate to 0.006, and improves braking response and control-scenario progress to 0.550 and 0.872, respectively. The data flywheel therefore substantially alleviates policy stalling while preserving a relatively high level of collision avoidance.
\begin{table}[t]
\centering
\caption{Open-loop results across three driving models. Bootstrap 95\% confidence intervals are reported in Appendix~E.}
\label{tab:main_results}

\definecolor{RFiveBlue}{HTML}{EAF2FB}

\newcommand{\TOneMax}[1]{\textbf{#1}}
\newcommand{\TOneVal}[2]{#1\,{\footnotesize ($#2\%$)}}

\small
\setlength{\tabcolsep}{2.8pt}
\renewcommand{\arraystretch}{0.92}

\begin{adjustbox}{max width=\columnwidth,center}
\begin{tabular}{@{}cccccccccc@{}}
\toprule
Model & Round & \textbf{CFR} $\uparrow$ & AD $\uparrow$ & BR $\uparrow$ & RS $\uparrow$ & RM $\uparrow$ & UBR $\downarrow$ & CSP $\uparrow$ & \(L_{\mathrm{val}}\downarrow\) \\
\midrule

\multirow[c]{6}{*}{\textbf{AutoVLA}} & Base & 0.261 & 0.100 & 0.090 & 0.984 & 0.985 & 0.000 & 0.751 & 1.249 \\
& R1 & 0.396 & 0.261 & 0.055 & 0.971 & 0.980 & 0.000 & 0.826 & \TOneVal{1.105}{-11.5} \\
& R2 & 0.519 & 0.431 & 0.110 & 0.973 & 1.000 & 0.005 & 0.933 & \TOneVal{1.036}{-17.1} \\
& R3 & 0.544 & 0.558 & 0.085 & 0.973 & 1.000 & 0.006 & 0.950 & \TOneVal{1.035}{-17.2} \\
& R4 & 0.578 & 0.581 & 0.115 & 0.973 & 1.000 & 0.006 & 0.906 & \TOneVal{1.054}{-15.6} \\
\rowcolor{RFiveBlue}\cellcolor{white} & R5 & 0.579 & 0.575 & 0.115 & 0.973 & 1.000 & 0.013 & 0.914 & \TOneVal{1.084}{-13.2} \\

\midrule

\multirow[c]{6}{*}{\textbf{ReCogDrive}} & Base & 0.390 & 0.297 & 0.045 & 0.731 & 0.995 & 0.001 & 0.884 & 0.366 \\
& R1 & 0.667 & 0.544 & 0.190 & 0.995 & 1.000 & 0.001 & 0.791 & \TOneVal{0.374}{+2.3} \\
& R2 & 0.708 & 0.508 & 0.175 & 0.952 & 1.000 & 0.001 & 0.676 & \TOneVal{0.382}{+4.2} \\
& R3 & 0.751 & 0.567 & 0.165 & 0.989 & 1.000 & 0.003 & 0.669 & \TOneVal{0.386}{+5.4} \\
& R4 & 0.746 & 0.536 & 0.140 & 1.000 & 1.000 & 0.001 & 0.680 & \TOneVal{0.371}{+1.3} \\
\rowcolor{RFiveBlue}\cellcolor{white} & R5 & 0.765 & 0.631 & 0.185 & 0.976 & 1.000 & 0.001 & 0.733 & \TOneVal{0.377}{+3.1} \\

\midrule

\multirow[c]{6}{*}{\textbf{LAW}} & Base & 0.967 & 1.000 & 0.035 & 0.925 & 0.185 & 1.000 & 0.113 & 0.633 \\
& R1 & 0.786 & 0.847 & 0.540 & 0.856 & 0.190 & 0.029 & 0.876 & \TOneVal{0.668}{+5.5} \\
& R2 & 0.785 & 0.861 & 0.540 & 0.859 & 0.180 & 0.005 & 0.867 & \TOneVal{0.650}{+2.7} \\
& R3 & 0.796 & 0.800 & 0.550 & 0.878 & 0.140 & 0.001 & 0.879 & \TOneVal{0.640}{+1.1} \\
& R4 & 0.795 & 0.883 & 0.535 & 0.902 & 0.115 & 0.011 & 0.840 & \TOneVal{0.639}{+0.9} \\
\rowcolor{RFiveBlue}\cellcolor{white} & R5 & 0.791 & 0.872 & 0.550 & 0.891 & 0.220 & 0.006 & 0.872 & \TOneVal{0.637}{+0.6} \\

\bottomrule
\end{tabular}
\end{adjustbox}
\end{table}

\noindent\textbf{Closed-loop performance.}
Compared with the open-loop models, two rounds of closed-loop post-training consistently improve the ability of all three models to complete the target scenarios (Table~\ref{tab:closed_loop_results}). The safe-passage rates of AutoVLA, ReCogDrive, and LAW increase from 0.227, 0.429, and 0.227 to 0.785, 0.859, and 0.423, respectively, corresponding to gains of 55.8, 43.0, and 19.6 percentage points. Their target-scenario progress also increases from 0.741, 0.632, and 0.632 to 0.873, 0.891, and 0.744, respectively. These joint improvements show that closed-loop post-training enables the models to make progress through the target scenarios and clear the conflict point safely without excessive slowing or stalling.

AutoVLA and LAW exhibit two distinct patterns. For AutoVLA, from the base model to the open-loop model, the pedestrian collision rate under closed-loop evaluation increases from 0.356 to 0.749, while the safe-passage rate decreases from 0.595 to 0.227. Closed-loop post-training reverses this degradation, reducing the pedestrian collision rate to 0.123 and increasing the safe-passage rate to 0.785, both surpassing the corresponding results of the base model. This result shows that improvements obtained through open-loop post-training do not necessarily transfer directly to sustained interaction, and that closed-loop post-training can address behavioral failures exposed during closed-loop execution. LAW follows a different pattern. Although its base model has no pedestrian collisions, its safe-passage rate is zero, and its progress in the target and paired control scenarios is only 0.040 and 0.041, respectively, indicating that its zero collision rate arises mainly from policy stalling. Post-training alleviates this behavior, allowing LAW to make progress and complete a subset of the target scenarios safely while maintaining a low final pedestrian collision rate of 0.018.

In the paired control scenarios, the three models achieve final no-event rates of 0.963, 1.000, and 1.000. Relative to the open-loop models, the control-scenario progress of ReCogDrive increases from 0.788 to 0.954, while that of AutoVLA and LAW changes from 0.978 and 0.985 to 0.915 and 0.962, respectively. Despite these model-dependent changes, all three models retain control-scenario progress above 0.9 and high no-event rates, indicating that the improvements in the target capability largely preserve normal behavior in the paired control scenarios.

\begin{table}[t]
\centering
\caption{Closed-loop results across three driving models.
Base denotes the pretrained model, OL denotes the checkpoint retained
after open-loop post-training, and CL R1 and CL R2 denote the first
and second rounds of closed-loop post-training, respectively.
$s$ denotes the target scenario, and $\bar{s}$ denotes its paired
control. Safe passage requires the ego vehicle to clear the conflict
point without any terminal event.}
\label{tab:closed_loop_results}

\definecolor{RFiveBlue}{HTML}{EAF2FB}

\footnotesize
\setlength{\tabcolsep}{2.8pt}
\renewcommand{\arraystretch}{0.92}

\begin{adjustbox}{max width=\columnwidth,center}
\begin{tabular}{@{}llccccc@{}}
\toprule
Model
& Stage
& \multicolumn{3}{c}{Target Scenario $s$}
& \multicolumn{2}{c}{Paired Control $\bar{s}$} \\
\cmidrule(lr){3-5}
\cmidrule(lr){6-7}
&
& PC $\downarrow$
& SP $\uparrow$
& P $\uparrow$
& NER $\uparrow$
& P $\uparrow$ \\
\midrule

\multirow[c]{4}{*}{\textbf{AutoVLA}}
& Base  & 0.356 & 0.595 & 0.827 & 0.982 & 0.948 \\
& OL    & 0.749 & 0.227 & 0.741 & 0.914 & 0.978 \\
& CL R1 & 0.135 & 0.687 & 0.793 & 0.963 & 0.857 \\
\rowcolor{RFiveBlue}\cellcolor{white}
& CL R2 & 0.123 & 0.785 & 0.873 & 0.963 & 0.915 \\

\midrule

\multirow[c]{4}{*}{\textbf{ReCogDrive}}
& Base  & 0.436 & 0.153 & 0.620 & 0.988 & 0.747 \\
& OL    & 0.098 & 0.429 & 0.632 & 1.000 & 0.788 \\
& CL R1 & 0.080 & 0.847 & 0.889 & 1.000 & 0.952 \\
\rowcolor{RFiveBlue}\cellcolor{white}
& CL R2 & 0.049 & 0.859 & 0.891 & 1.000 & 0.954 \\

\midrule

\multirow[c]{4}{*}{\textbf{LAW}}
& Base  & 0.000 & 0.000 & 0.040 & 1.000 & 0.041 \\
& OL    & 0.000 & 0.227 & 0.632 & 0.994 & 0.985 \\
& CL R1 & 0.025 & 0.368 & 0.702 & 1.000 & 0.958 \\
\rowcolor{RFiveBlue}\cellcolor{white}
& CL R2 & 0.018 & 0.423 & 0.744 & 1.000 & 0.962 \\

\bottomrule
\end{tabular}
\end{adjustbox}
\end{table}

\subsection{Sequential Acquisition of a Second Capability}

To examine whether the data flywheel can acquire a new capability while retaining an existing one, we post-train AutoVLA and ReCogDrive for four rounds on stop-sign-controlled intersections, starting from either the base model or the occluded-pedestrian checkpoint. The primary metric is the Stop-before-Line Rate (SLR), which measures sustained stops within 5\,m before the stop line. Other metrics include Yielding Rate (YR), Resuming Motion (RM), Unnecessary Braking Rate (UBR), Early Stopping Rate (ESR), and Conflict-Vehicle Collision Rate (CVCR). UBR is evaluated when the stop sign applies to another approach. We also report the original-task validation error ($L_{\mathrm{val}}$) and, for sequential training, the collision-free rate on the previously acquired occluded-pedestrian capability ($\mathrm{CFR}_{\mathrm{prev}}$). Appendix~B details event generation, post-training, and the evaluation protocol.

Table~\ref{tab:sequential_acquisition} reports the results after four
rounds. For AutoVLA, the stop-before-line rate increases from 0.215 to
0.599 with sequential training and from 0.100 to 0.897 when training
from the base checkpoint. Its pedestrian collision-free rate decreases
from 0.579 to 0.569 during sequential training. For ReCogDrive, the stop-before-line rate increases from 0.003 to 0.358 with sequential
training and from 0.000 to 0.667 when training from the base checkpoint.
Its pedestrian collision-free rate decreases from 0.765 to 0.733.
Both models largely retain the previous capability, with pedestrian collision-free rates decreasing by 0.010 and 0.032. The stop-before-line rate improves under both initializations, more slowly when training starts from the pedestrian checkpoints.

\begin{table}[t]
\centering
\caption{Sequential acquisition of the stop-sign-controlled intersection
capability on AutoVLA and ReCogDrive. Base and Ped. denote initialization
from the base model and the occluded-pedestrian checkpoint, respectively.}
\label{tab:sequential_acquisition}

\definecolor{RFiveBlue}{HTML}{EAF2FB}

\newcommand{\TThreeVal}[2]{#1\,{\footnotesize ($#2\%$)}}

\small
\setlength{\tabcolsep}{2.8pt}
\renewcommand{\arraystretch}{0.92}

\begin{adjustbox}{max width=\columnwidth,center}
\begin{tabular}{@{}lccccccccc@{}}
\toprule
Model
& Round
& \textbf{SLR} $\uparrow$
& YR $\uparrow$
& RM $\uparrow$
& UBR $\downarrow$
& ESR $\downarrow$
& CVCR $\downarrow$
& $L_{\mathrm{val}}\downarrow$
& $\mathrm{CFR}_{\mathrm{prev}}\uparrow$ \\
\midrule

\multicolumn{10}{@{}l}{\textbf{AutoVLA}} \\

\multirow[c]{5}{*}{Base}
& Init. & 0.100 & 0.461 & 1.000 & 0.064 & 0.000 & 0.101
& 1.249 & -- \\
& R1 & 0.513 & 0.833 & 1.000 & 0.032 & 0.000 & 0.005
& \TThreeVal{1.048}{-16.1} & -- \\
& R2 & 0.699 & 1.000 & 0.963 & 0.000 & 0.000 & 0.025
& \TThreeVal{1.166}{-6.6} & -- \\
& R3 & 0.791 & 1.000 & 1.000 & 0.038 & 0.000 & 0.015
& \TThreeVal{1.082}{-13.4} & -- \\
\rowcolor{RFiveBlue}\cellcolor{white}
& R4 & 0.897 & 0.984 & 1.000 & 0.036 & 0.000 & 0.000
& \TThreeVal{1.084}{-13.2} & -- \\

\addlinespace

\multirow[c]{5}{*}{Ped.}
& Init. & 0.215 & 0.527 & 0.995 & 0.107 & 0.000 & 0.097
& \TThreeVal{1.084}{-13.2} & 0.579 \\
& R1 & 0.324 & 1.000 & 0.979 & 0.028 & 0.000 & 0.071
& \TThreeVal{1.089}{-12.8} & 0.568 \\
& R2 & 0.360 & 0.976 & 0.969 & 0.060 & 0.000 & 0.017
& \TThreeVal{1.116}{-10.6} & 0.584 \\
& R3 & 0.498 & 0.996 & 0.953 & 0.036 & 0.000 & 0.018
& \TThreeVal{1.105}{-11.5} & 0.565 \\
\rowcolor{RFiveBlue}\cellcolor{white}
& R4 & 0.599 & 1.000 & 0.958 & 0.041 & 0.000 & 0.070
& \TThreeVal{1.089}{-12.8} & 0.569 \\

\midrule

\multicolumn{10}{@{}l}{\textbf{ReCogDrive}} \\

\multirow[c]{5}{*}{Base}
& Init. & 0.000 & 0.588 & 1.000 & 0.000 & 0.000 & 0.022
& 0.366 & -- \\
& R1 & 0.422 & 0.604 & 1.000 & 0.000 & 0.000 & 0.000
& \TThreeVal{0.384}{+4.9} & -- \\
& R2 & 0.659 & 0.580 & 1.000 & 0.000 & 0.000 & 0.000
& \TThreeVal{0.376}{+2.7} & -- \\
& R3 & 0.457 & 0.571 & 1.000 & 0.000 & 0.000 & 0.000
& \TThreeVal{0.370}{+1.1} & -- \\
\rowcolor{RFiveBlue}\cellcolor{white}
& R4 & 0.667 & 0.592 & 1.000 & 0.000 & 0.000 & 0.000
& \TThreeVal{0.372}{+1.6} & -- \\

\addlinespace

\multirow[c]{5}{*}{Ped.}
& Init. & 0.003 & 0.482 & 1.000 & 0.006 & 0.000 & 0.000
& \TThreeVal{0.377}{+3.1} & 0.765 \\
& R1 & 0.249 & 0.625 & 1.000 & 0.000 & 0.000 & 0.000
& \TThreeVal{0.378}{+3.3} & 0.753 \\
& R2 & 0.331 & 0.559 & 1.000 & 0.000 & 0.000 & 0.000
& \TThreeVal{0.376}{+2.7} & 0.669 \\
& R3 & 0.364 & 0.592 & 1.000 & 0.000 & 0.000 & 0.000
& \TThreeVal{0.379}{+3.6} & 0.700 \\
\rowcolor{RFiveBlue}\cellcolor{white}
& R4 & 0.358 & 0.584 & 1.000 & 0.000 & 0.000 & 0.000
& \TThreeVal{0.384}{+4.9} & 0.733 \\

\bottomrule
\end{tabular}
\end{adjustbox}
\end{table}

\subsection{Ablation Study}

\noindent\textbf{Failure feedback and multi-round post-training.}
We evaluate the effects of data generation quotas and multi-round post-training on ReCogDrive. To isolate the effect of failure feedback, four multi-round variants start from the same model retained after the first round and subsequently adopt dynamically updated failure-feedback-driven quotas, quotas based on a frozen failure map (FF), uniform quotas (UQ), or random sampling (Rand). To isolate the effect of multi-round post-training, we further compare multi-round (MR) and single-stage post-training (SS) under uniform quotas and random sampling, using identical generated data within each comparison. All settings are matched by cumulative training steps. At each training budget, we select the candidate checkpoint with the highest primary metric score on the development set among those satisfying the constraints. Results are shown in Figure~\ref{fig:budget}, with detailed numerical results provided in Appendix~C.

\begin{figure}[t]
  \centering
  \includegraphics[width=0.85\linewidth]{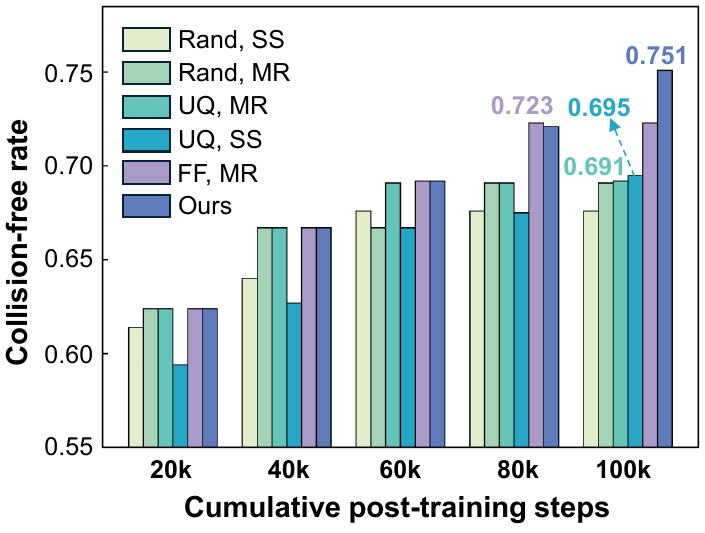} 
  \caption{Sealed-test critical-phase collision-free rates of ReCogDrive across training budgets for six schemes.}
  \label{fig:budget}
\end{figure}

The four multi-round variants share the first round and therefore achieve identical performance at 40k steps. Their trajectories subsequently diverge: UQ and Rand plateau at approximately 0.69, whereas the full flywheel reaches 0.751 at 100k steps, outperforming the two MR baselines by 6.0 percentage points. FF reaches 0.723 at 80k steps but shows no further improvement at 100k, leaving a 2.8-percentage-point gap to the full flywheel (a collision rate of 0.277 versus 0.249, a 10.1\% relative reduction). Without failure feedback, MR and SS perform comparably at 100k steps: the two UQ settings achieve 0.691 and 0.695, respectively, while the Rand settings differ by only 1.5 percentage points. These results suggest that failure-guided data allocation is beneficial even with a frozen map, while updating the map across rounds yields additional gains beyond optimization or the MR schedule alone. The full set of metrics at the 100k-step budget is given in Appendix~C.

\noindent\textbf{Paired control scenario samples.}
We remove the paired control scenario samples from the ReCogDrive training pool while keeping all other settings unchanged. Their removal increases the primary metric from 0.751 to 0.944 (Table~\ref{tab:paired_control_ablation}). However, the unnecessary braking rate rises from 0.003 to 0.050, control-scenario progress decreases from 0.669 to 0.224, and braking response drops from 0.165 to 0.035. These results indicate that the apparent improvement in the primary metric arises from indiscriminate braking. Paired control scenario samples suppress this overly conservative degenerate solution.

\begin{table}[t]
\centering
\caption{Ablation of paired control scenario samples on ReCogDrive.}
\label{tab:paired_control_ablation}

\definecolor{HighlightBlue}{HTML}{EAF2FB}

\newcommand{\TFourVal}[2]{#1\,{\footnotesize ($#2\%$)}}

\small
\setlength{\tabcolsep}{2.8pt}
\renewcommand{\arraystretch}{0.92}

\begin{adjustbox}{max width=\columnwidth,center}
\begin{tabular}{@{}lcccccccc@{}}
\toprule
Training Data
& \textbf{CFR} $\uparrow$
& AD $\uparrow$
& BR $\uparrow$
& RS $\uparrow$
& RM $\uparrow$
& UBR $\downarrow$
& CSP $\uparrow$
& $L_{\mathrm{val}}\downarrow$ \\
\midrule

No paired controls
& 0.944
& 0.497
& 0.035
& 0.973
& 1.000
& 0.050
& 0.224
& \TFourVal{0.379}{+3.6} \\

Ours
& 0.751
& 0.567
& 0.165
& 0.989
& 1.000
& 0.003
& 0.669
& \TFourVal{0.386}{+5.4} \\

\bottomrule
\end{tabular}
\end{adjustbox}
\end{table}

Additional ablations of original-task data replay, the model acceptance rule, and validity checks are provided in Appendix C.

\subsection{Real-Vehicle Evaluation}
\label{sec:real_vehicle}

We directly deploy the ReCogDrive base model and the final checkpoints from open-loop and closed-loop post-training on a full-scale vehicle, without additional real-world training. 
The target and paired control scenarios are reproduced on a closed test track, with a pedestrian dummy used in the target scenarios. 
In closed loop, each checkpoint controls the vehicle speed in five target and five paired control trials. 
In open loop, the rule-based expert drives the vehicle, and each checkpoint is queried on the recorded camera observations using the metrics in Table~\ref{tab:main_results}. 
Experimental details and the open-loop results of AutoVLA and LAW on the recorded observations are provided in Appendix~D.

In the closed-loop trials, the number of safe passages increases from 1 of 5 for the base model to 4 of 5 for the open-loop checkpoint (Table~\ref{tab:real_vehicle}). 
At pedestrian emergence, the open-loop checkpoint travels at a median speed of 2.87\,m/s, compared with 3.98\,m/s for the base model, which leaves 3.6\,s versus 1.9\,s before the conflict point. 
After emergence, it reduces its speed by a further 1.71\,m/s, compared with 0.96\,m/s for the base model, although no checkpoint brings the vehicle to a full stop. 
All paired control trials finish without a terminal event, and the open-loop checkpoint maintains a minimum speed comparable to that of the base model in these trials (3.91 and 3.62\,m/s). 

On the recorded observations, open-loop post-training increases the critical-phase collision-free rate from 0.781 to 0.989 and anticipatory deceleration from 0.320 to 0.369, whereas the braking response remains similar (0.512 and 0.492). 
This improvement is absent for the closed-loop checkpoint, which travels at 4.36\,m/s at emergence and achieves 2 of 5 safe passages and a collision-free rate of 0.645 on the recorded observations. 
A possible reason is that closed-loop post-training emphasizes the first predicted waypoint over the rest of the plan (Appendix~D).
For the open-loop checkpoint, both evaluations are consistent with the simulation results in Table~\ref{tab:main_results}, where the improvement arises mainly from anticipatory deceleration and the braking response after pedestrian emergence remains limited. 

\begin{table}[t]
    \centering
    \caption{
        Real-vehicle evaluation of ReCogDrive.
        SP is the number of five closed-loop target trials completed without a braking intervention by the safety driver due to collision risk.
        $v_{\mathrm{e}}$ (m/s), $t_{\mathrm{e}}$ (s), and $\Delta v_{\mathrm{e}}$ (m/s) are the median speed at pedestrian emergence, time to the conflict point at emergence, and speed reduction after emergence.
        $v_{\min}$ (m/s) is the median minimum speed in the five paired control trials.
        OL and CL denote the final checkpoints from open-loop and closed-loop post-training.
    }
    \label{tab:real_vehicle}
    \footnotesize
    \setlength{\tabcolsep}{1.5pt}
    \renewcommand{\arraystretch}{1.1}
    \begin{tabular}{@{}lcccccccccc@{}}
        \toprule
        & \multicolumn{4}{c}{Closed loop, $s$} & $\bar{s}$ & \multicolumn{5}{c}{Open loop} \\
        \cmidrule(lr){2-5}\cmidrule(lr){6-6}\cmidrule(l){7-11}
        Model & \textbf{SP} $\uparrow$ & $v_{\mathrm{e}}$ & $t_{\mathrm{e}}$ & $\Delta v_{\mathrm{e}}$ & $v_{\min}$ & \textbf{CFR} $\uparrow$ & AD $\uparrow$ & BR $\uparrow$ & UBR $\downarrow$ & CSP \\
        \midrule
        Base & 1/5 & 3.98 & 1.9 & 0.96 & 3.62 & 0.781 & 0.320 & 0.512 & 0.000 & 0.832 \\
        OL   & 4/5 & 2.87 & 3.6 & 1.71 & 3.91 & 0.989 & 0.369 & 0.492 & 0.000 & 0.722 \\
        CL   & 2/5 & 4.36 & 1.9 & 0.77 & 3.38 & 0.645 & 0.371 & 0.440 & 0.000 & 0.864 \\
        \bottomrule
    \end{tabular}
\end{table}

\section{Conclusion}
\label{sec:conclusion}
This paper presents a failure-driven data flywheel for post-training VLA models that constructs target scenarios absent from the available data, evaluates individual behavioral requirements, and assigns data generation quotas from a failure map updated after each round.
For two target capabilities in autonomous driving, the framework improves the target capability while largely retaining existing ones, whereas fine-grained evaluation shows that the braking response after pedestrian emergence remains limited.
The current validation is limited to driving and a rule-based expert, and improving the braking response and extending the framework to other platforms are left for future work.

\section*{LLM Usage Statement}
We used Claude (Anthropic) and ChatGPT (OpenAI) for two purposes.
First, they assisted in drafting and polishing the text of the paper.
Second, they assisted in writing, debugging, and running the code for our experiments. The authors reviewed all content and take full responsibility for it.

{
    \small
    \bibliographystyle{ieeenat_fullname}
    \bibliography{main}
}

\twocolumn[{\centering\Large\textbf{Appendix}\par\vspace{1.0em}}]
\appendix
\counterwithin{equation}{section}
\counterwithin{table}{section}
\counterwithin{figure}{section}
\renewcommand{\thetable}{\thesection-\arabic{table}}
\renewcommand{\thefigure}{\thesection-\arabic{figure}}
\section{Overview and Notation}
\label{sec:appendix_overview}

This appendix is organized as follows.
Appendix~A summarizes the notation used in the main paper and the appendices (Table~\ref{tab:notation}).
Appendix~B provides the implementation details of scenario construction, the asset library, the rule-based expert, open-loop and closed-loop post-training, the capability metrics and constraints, and the stop-sign yielding instantiation.
Appendix~C reports the numerical results of Figure~3, the full set of metrics at the 100k-step budget, and additional ablations of original-task data replay, the model acceptance rule, and validity checks.
Appendix~D describes the real-vehicle experiments, reports the open-loop results of AutoVLA and LAW on the recorded observations, and analyzes the closed-loop checkpoint.
Appendix~E reports the bootstrap 95\% confidence intervals for Table~1.

\begin{table*}[t]
\centering
\caption{Notation.
The first five groups list symbols defined in the main paper, and the last group lists symbols introduced in Appendix~B.
The last column gives the section, equation, or table in which each symbol is defined.}
\label{tab:notation}
\footnotesize
\setlength{\tabcolsep}{6pt}
\renewcommand{\arraystretch}{1.0}
\begin{tabular}{@{}l p{0.68\textwidth} l@{}}
\toprule
Symbol & Meaning & Defined in \\
\midrule
\multicolumn{3}{@{}l}{\textit{Capability and scenario}} \\
$c=(\mathcal{T},\Theta,\pi^\star,\mathcal{F})$ & Target capability, with scenario template $\mathcal{T}$, space of scenario parameters $\Theta$, and expert $\pi^\star$ & Sec.~3.1 \\
$\mathcal{F}=(F_1,\dots,F_k)$ & Tuple of capability metrics, with $F_1$ as the primary metric & Sec.~3.1 \\
$g$ & Base scene & Sec.~3.1 \\
$\theta$, $\Theta_{\mathrm{feas}}(g)$ & Scenario parameters and the feasible configurations for base scene $g$ & Sec.~3.1 \\
$s=\mathcal{T}(g,\theta)$, $\bar{s}$ & Target scenario and its paired control scenario & Sec.~3.1 \\
$d(\theta)$ & Target time to collision (TTC) at first pedestrian visibility & Sec.~3.2 \\
$\tau^\star=\pi^\star(s)$, $\bar{\tau}^\star=\pi^\star(\bar{s})$ & Expert demonstrations for the target scenario and the paired control scenario & Sec.~3.3 \\
\addlinespace
\multicolumn{3}{@{}l}{\textit{Event timing for occluded-pedestrian emergence}} \\
$v_{\mathrm{safe}}$, $L$ & Speed of the nominal ego approach and ego-vehicle length & Eq.~(1) \\
$u_{\mathrm{conf}}$, $u_{\mathrm{rev}}$ & Conflict-point position along the ego path and nominal ego position at emergence & Eq.~(1) \\
$t^{\mathrm{nom}}_{\mathrm{rev}}$, $t_{\mathrm{depart}}$ & Time at which the nominal ego motion reaches $u_{\mathrm{rev}}$ and pedestrian departure time & Eq.~(1) \\
$\ell_{\mathrm{vis}}$, $v_p$ & Distance traveled by the pedestrian before becoming visible and pedestrian speed & Eq.~(1) \\
\addlinespace
\multicolumn{3}{@{}l}{\textit{Post-training}} \\
$\pi_0$, $r$ & Pretrained model and post-training round & Sec.~3.4 \\
$\mathcal{D}^{\mathrm{gen}}_r$ & Generated-data pool in round $r$ & Sec.~3.4 \\
$p^{\mathrm{train}}_r$, $p^{\mathrm{gen}}_r$, $p^{\mathrm{orig}}$ & Training distribution in round $r$ and sampling distributions over $\mathcal{D}^{\mathrm{gen}}_r$ and over the original-task data & Eq.~(2) \\
$\rho$ & Original-task replay proportion & Eq.~(2) \\
$L_{\mathrm{val}}$ & Original-task validation error & Sec.~4.1 \\
\addlinespace
\multicolumn{3}{@{}l}{\textit{Capability evaluation and generation quotas}} \\
$t$, $o_t$, $\hat{\tau}_t=\pi(o_t)$ & Query time, observation, and the future trajectory predicted by the model $\pi$ & Sec.~3.5 \\
$\mathcal{S}^{\mathrm{dev}}$ & Fixed development set & Sec.~3.5 \\
$\mathcal{B}$, $b$ & Set of task-condition and difficulty-level bins and one of its bins & Sec.~3.5 \\
$\phi(g,\theta)$ & Grouping function that assigns each target scenario to a bin & Sec.~3.5 \\
$m_r(b)$ & Failure rate of the retained checkpoint within bin $b$ after round $r$ & Sec.~3.5 \\
$q_{r+1}(b)$ & Proportion of the next round's generation budget allocated to bin $b$ & Eq.~(3) \\
$\epsilon_q$ & Allocation-weight floor & Eq.~(3) \\
\addlinespace
\multicolumn{3}{@{}l}{\textit{Real-vehicle evaluation}} \\
$v_{\mathrm{e}}$, $t_{\mathrm{e}}$, $\Delta v_{\mathrm{e}}$ & Median speed at pedestrian emergence, time to the conflict point at emergence, and speed reduction after emergence & Table~5 \\
$v_{\min}$ & Median minimum speed in the paired control trials & Table~5 \\
\addlinespace
\multicolumn{3}{@{}l}{\textit{Symbols introduced in Appendix~B}} \\
$a_{\mathrm{req}}(d)$ & Required deceleration for a target TTC of $d$ & B.1 \\
$a_{\mathrm{dec}}$, $a_{\mathrm{acc}}$ & Deceleration of the expert during the approach and its acceleration when resuming motion & B.3 \\
$\tau_r$, $d_{\mathrm{stop}}$ & Reaction delay and stopping margin of the expert & B.3 \\
$t_{\mathrm{ease}}$, $t_{\mathrm{done}}$ & Times at which the expert begins anticipatory deceleration and reaches $v_{\mathrm{safe}}$ & B.6 \\
$t_{\mathrm{rev}}$, $t_{\mathrm{brake}}$ & Pedestrian emergence time and the time at which the expert begins braking & B.6 \\
$t_{\mathrm{stop}}$, $t_{\mathrm{resume}}$, $\bar{t}_{\mathrm{resume}}$ & Times at which the expert comes to a stop and resumes motion, and the resumption time in the paired control & B.6 \\
$v^{(1)}_t$, $d_t(h)$, $\ell_t$ & Mean speed over the first $0.5$\,s segment, displacement over the first $h$ seconds, and arc length of the predicted trajectory & B.6 \\
$\bar{v}^{\star(1)}_t$, $\bar{\ell}^{\star}_t$ & First-segment mean speed and arc length of the expert trajectory in the paired control & B.6 \\
$T_e$ & End of the closed-loop scoring window & B.7 \\
$\Delta t$ & Arrival offset of the first conflict vehicle relative to the time at which the ego vehicle comes to a stop & B.8 \\
\bottomrule
\end{tabular}
\end{table*}

\section{Implementation Details}
\label{sec:appendix_implementation}

\subsection{Scenario Construction for Occluded-Pedestrian Emergence}
\label{sec:appendix_scenario}

\noindent\textbf{Scenario template.}
We place a parked vehicle near the curb on either side of the recorded ego route to serve as an occluder.
A pedestrian starts from the end of the occluder farther from the approaching ego vehicle and walks at a constant speed along a straight path across the ego route.
The intersection of the pedestrian path and the ego route defines the conflict point.
The scenario parameters $\theta$ specify the conflict-point location, the pedestrian and occluder assets and their placements, the pedestrian walking speed, and the target time to collision (TTC) $d(\theta)$ at first pedestrian visibility.
The paired control scenario $\bar{s}$ retains the same base scene, occluder placement, and ego approach trajectory as the target scenario $s$, but omits the pedestrian.

\noindent\textbf{Pedestrian emergence.}
The emergence frame is the first frame in which the pedestrian is visible to at least one of the three forward-facing cameras.
The pedestrian departure time is solved backward from the target TTC using Eq.~(1).
The pedestrian emerges only after the ego front bumper has passed the end of the occluder facing the approaching vehicle.
At that point, the ego speed has been reduced to $v_{\mathrm{safe}} = 4$\,m/s.
The target and paired control scenarios therefore yield identical observations before emergence.

\noindent\textbf{Feasible parameter space.}
For each base scene $g$, we enumerate candidate configurations over 41 keyframes and retain those satisfying the geometric, temporal, and braking constraints that define $\Theta_{\mathrm{feas}}(g)$.
The occluder must be parked near a curb, have at least 78\% of its footprint within the drivable area, and be at least as tall as the sampled pedestrian.
It must also maintain lateral clearance between its outer boundary and the ego-route centerline without intersecting static objects or recorded vehicle trajectories.
The pedestrian path must be obstacle-free, with at least 75\% of its length within traversable areas, and must remain at least $2.5$\,m from recorded moving vehicles.
We further require the pedestrian to remain occluded for at least $0.7$\,s before emergence, the recorded ego speed at the conflict time to be at least $2$\,m/s, the estimated pedestrian height at emergence to be at least 50 pixels, and the deceleration required by the expert trajectory to be no greater than $8$\,m/s$^2$.

\noindent\textbf{Difficulty bins.}
We characterize event difficulty by the nominal braking demand associated with the target TTC.
Suppose that the ego vehicle travels at $4$\,m/s at emergence, maintains this speed during a $0.5$\,s reaction delay, and then brakes at a constant rate until its front bumper stops $1.0$\,m before the pedestrian.
The required deceleration is
\begin{equation}
a_{\mathrm{req}}(d) = \frac{8}{4d-3}\ \mathrm{m/s^2},
\label{eq:a_req}
\end{equation}
where $d = d(\theta)$ is measured in seconds.
The four TTC bins $[1.10, 1.15)$, $[1.15, 1.35)$, $[1.35, 1.60)$, and $[1.60, 2.00]$\,s correspond to nominal deceleration ranges of approximately $5.0$--$5.7$, $3.3$--$5.0$, $2.35$--$3.3$, and $1.6$--$2.35$\,m/s$^2$, respectively.

\noindent\textbf{Evaluation-set sampling.}
Each TTC bin in both the development and sealed test sets contains 40 adult events and 10 child events, with TTC sampled uniformly within the bin.
Each set therefore contains 200 target scenarios and their paired controls.
Table~\ref{tab:set_parameters} summarizes the main parameters of the two sets.

\begin{table}[t]
\centering
\caption{Main parameters of the sealed test and development sets.
Slash-separated triplets report the minimum, median, and maximum, respectively.}
\label{tab:set_parameters}
\small
\setlength{\tabcolsep}{4pt}
\renewcommand{\arraystretch}{0.95}
\begin{adjustbox}{max width=\columnwidth,center}
\begin{tabular}{@{}lcc@{}}
\toprule
Parameter & Sealed Test Set & Development Set \\
\midrule
Target TTC & 50 per bin & 50 per bin \\
Pedestrian walking speed (m/s) & 1.27 / 2.80 / 2.80 & 1.27 / 2.20 / 2.80 \\
Occluder length (m) & 3.66 / 6.49 / 13.65 & 2.77 / 6.49 / 13.65 \\
Occluder side, right / left & 150 / 50 & 131 / 69 \\
Pedestrian group, adult / child & 160 / 40 & 160 / 40 \\
\bottomrule
\end{tabular}
\end{adjustbox}
\end{table}

\subsection{Asset Library and Screening}
\label{sec:appendix_assets}

\noindent\textbf{Pedestrian assets.}
The pedestrian asset library contains 2,545 adult models and eight child models.
The adult assets comprise 2,445 human scans from THuman2.1~\cite{tao2021function4d} and 100 from HSRD~\cite{HSRD-100_2025}.
Each inserted adult is animated with walking poses of real pedestrians, which are taken from the pedestrian poses obtained by OmniRe~\cite{chen2025omnire} when reconstructing the nuScenes~\cite{caesar2020nuscenes} scenes.
The child assets are rigged models spanning four heights of 1.10, 1.20, 1.30, and 1.40\,m, with two assets at each height.
Three are based on Microsoft Rocketbox~\cite{gonzalez2020rocketbox}, and five are generated using MakeHuman~\cite{bastioni2008ideas}.

\noindent\textbf{Occluder assets and screening.}
The candidate library contains 626 vehicle assets, including 562 scanned vehicles from 3DRealCar~\cite{du20253drealcar} and 64 mesh models from Objaverse~\cite{deitke2023objaverse}.
We retain an asset only if the lower boundary of its opaque body region lies no more than $0.5$\,m above the ground, its 90th-percentile hole ratio is at most 0.05, and its fragmentation ratio is at most 0.01.
A total of 226 vehicles satisfy these criteria, including 179 from 3DRealCar and 47 from Objaverse.
For each event, we further require the opaque region of the selected vehicle to extend to at least the height of the sampled pedestrian.

\noindent\textbf{Asset usage in the evaluation sets.}
Table~\ref{tab:asset_usage} reports the number of target events associated with each asset source.

\begin{table}[t]
\centering
\caption{Asset usage in the sealed test and development sets.
Each set contains 200 target events.
}
\label{tab:asset_usage}
\small
\setlength{\tabcolsep}{4pt}
\renewcommand{\arraystretch}{0.95}
\begin{adjustbox}{max width=\columnwidth,center}
\begin{tabular}{@{}llcc@{}}
\toprule
Asset Class & Source or Type & Sealed Test & Development \\
\midrule
Pedestrian & THuman2.1 adult & 157 (78.5\%) & 149 (74.5\%) \\
Pedestrian & HSRD adult & 3 (1.5\%) & 11 (5.5\%) \\
Pedestrian & Child model & 40 (20.0\%) & 40 (20.0\%) \\
\midrule
Occluder & Objaverse & 171 (85.5\%) & 160 (80.0\%) \\
Occluder & 3DRealCar & 29 (14.5\%) & 40 (20.0\%) \\
\bottomrule
\end{tabular}
\end{adjustbox}
\end{table}

\subsection{Rule-Based Expert for Occluded-Pedestrian Emergence}
\label{sec:appendix_expert}

\noindent\textbf{Approach and anticipatory deceleration.}
The rule-based expert follows the recorded ego route.
The ego vehicle initially travels at the recorded speed and then decelerates at $a_{\mathrm{dec}} = 2$\,m/s$^2$ to $v_{\mathrm{safe}} = 4$\,m/s before its front bumper reaches the near end of the occluder.
It subsequently maintains this speed while approaching the occluded region.
The target scenario $s$ and its paired control $\bar{s}$ share this approach profile, yielding identical expert trajectories before pedestrian emergence.

\noindent\textbf{Braking response after emergence.}
Following first pedestrian visibility, the expert maintains $v_{\mathrm{safe}}$ for a reaction delay of $\tau_r = 0.5$\,s and then applies a constant deceleration determined by the target TTC, capped at $8$\,m/s$^2$.
The nominal stopping position places the ego front bumper $d_{\mathrm{stop}} = 1.0$\,m before the boundary of the pedestrian's circular footprint, whose radius is $0.3$\,m.

\noindent\textbf{Remaining stopped and resuming motion.}
The ego vehicle remains stopped until the pedestrian has moved $1.5$\,m beyond the ego-route centerline, followed by an additional waiting period of $0.5$\,s.
The expert then accelerates at $a_{\mathrm{acc}} = 1.5$\,m/s$^2$ until it recovers the recorded speed.
If the pedestrian clears the conflict region before the ego vehicle comes to a complete stop, the stopping phase is omitted and the expert transitions directly to resuming motion.

\noindent\textbf{Paired control scenario.}
For the paired control $\bar{s}$, the expert follows the same recorded-speed and anticipatory-deceleration segments as in the target scenario $s$.
It then maintains $v_{\mathrm{safe}}$ through the conflict point and accelerates at $1.5$\,m/s$^2$ until the recorded speed is recovered.

\subsection{Open-Loop Post-Training Protocol}
\label{sec:appendix_open_loop}

\noindent\textbf{Scenario budget and generation quotas.}
The first round generates 500 new target scenarios from 100 training base scenes.
Each base scene contributes four adult scenarios and one child scenario.
The adult scenarios are distributed uniformly across the four TTC bins.
From the second round onward, each round generates 400 new target scenarios, of which 20\% contain child pedestrians.
Following Eq.~(3), the generation weight for bin $b$ in round $r$ is proportional to $\max\{m_{r-1}(b), 0.05\}$, where $m_{r-1}(b)$ is the event-level failure rate of the preceding retained checkpoint on bin $b$ of the development set.

\noindent\textbf{Generated-data pool.}
The generated-data pool for round $r$, denoted by $\mathcal{D}_r^{\mathrm{gen}}$, contains supervision samples from the current round's target scenarios and paired control scenarios, together with samples rendered from the same base scenes without any inserted event.
Paired control samples account for approximately 30\% of the pool.
Samples without an inserted event use the recorded trajectory as the supervision target and account for at most 15\% of this pool.
From the second round onward, each selected event from a preceding round contributes one target-scenario sample.
These previous-round samples account for at most 10\% of the final generated-data pool.

\noindent\textbf{Original-task replay.}
Following Eq.~(2), the generated-data pool is mixed with independently sampled nuScenes original-task data using an original-task replay proportion of $\rho = 0.8$.

\noindent\textbf{Training configurations.}
The first round is initialized from the base checkpoint of each model, and each subsequent round is initialized from the checkpoint retained in the preceding round.
Each model retains its original training objective.
Table~\ref{tab:ol_configs} summarizes the model-specific configurations.

\begin{table*}[t]
\centering
\caption{Open-loop post-training configurations. The learning rate remains constant after warmup.}
\label{tab:ol_configs}
\footnotesize
\begin{tabular}{@{}l p{0.24\textwidth} p{0.27\textwidth} p{0.28\textwidth}@{}}
\toprule
Configuration & AutoVLA & ReCogDrive & LAW \\
\midrule
Base checkpoint & Official weights & Planner trained on the nuScenes training split & Three-camera model trained on the nuScenes training split \\
Updated parameters & Language model, with the vision encoder frozen & Diffusion planner, with the vision--language model frozen & All parameters, with a $0.1\times$ learning rate for the image backbone \\
Optimizer & AdamW & AdamW & AdamW \\
Learning rate & $1\times10^{-5}$ & $1\times10^{-5}$ & $5\times10^{-6}$ \\
Weight decay & $0.01$ & $1\times10^{-4}$ & $0.01$ \\
Warmup steps & 100 & 100 & 800 \\
Effective batch size & 12 & 16 & 8 \\
\bottomrule
\end{tabular}
\end{table*}

\noindent\textbf{Within-round early stopping and between-round acceptance.}
After each epoch, the resulting checkpoint is evaluated on the fixed development set, and its original-task validation error $L_{\mathrm{val}}$ is computed on the nuScenes validation split.
An epoch is considered to make progress if it improves the best development-set primary score observed in the current round by at least one percentage point.
Once an eligible checkpoint has been obtained, training stops after four consecutive epochs without such an improvement.
The round candidate is the checkpoint with the highest primary score on the development set among those satisfying the constraints of Appendix~B.6.
The candidate is accepted only when it achieves a strictly higher primary score than the incumbent checkpoint, and the incumbent is retained otherwise.
If the incumbent violates the constraints, the candidate replaces it directly.
The failure map for the next round is computed from the checkpoint retained in the current round, and the sealed test set is used only for final reporting.

\subsection{Query Times and Model-Specific Inputs and Outputs}
\label{sec:appendix_queries}

\noindent\textbf{Nominal query times.}
We select nominal query times along the expert trajectory to cover four behavioral stages: approach, pedestrian emergence and braking, remaining stopped, and resuming motion.
Each target scenario contains up to 12 nominal query times, and each paired control scenario contains four.
Table~\ref{tab:query_times} lists these query times and their corresponding metrics.
All query times are aligned to the $10$\,Hz rendering grid.
A query is discarded if its current frame or any required historical frame is unavailable.

\noindent\textbf{Supervision samples and model adaptation.}
Each supervision sample consists of the model input at query time $t$ and the corresponding future trajectory.
Target and paired control scenarios use their respective expert trajectories as supervision.
Samples from scenes without an inserted event and samples from original-task replay use the recorded trajectory.
For each model, we extract the appropriate prediction horizon from the same reference trajectory and convert it into the model's native coordinate system and output representation (Table~\ref{tab:model_io}).
All three models use a waypoint interval of $0.5$\,s.

\noindent\textbf{Navigation commands.}
Navigation commands follow the native convention of each model.
AutoVLA~\cite{zhou2026autovla} assigns one command to each event from the route covered by the expert trajectory, with the same command used for the target and paired control scenarios.
ReCogDrive~\cite{ICLR2026_ff7bf601} generates a command at each query time from the lateral offset of the nuScenes map route $20$\,m ahead.
LAW~\cite{li2025enhancing} follows the VAD~\cite{jiang2023vad} convention and derives the command from the lateral offset of the endpoint of the $3$\,s supervision trajectory.

\begin{table*}[t]
\centering
\caption{Nominal query times and their corresponding open-loop metrics. All nominal times are aligned to the $10$\,Hz rendering grid. For short braking intervals, a single braking midpoint replaces the one-third and two-thirds braking queries.}
\label{tab:query_times}
\footnotesize
\begin{tabular}{@{}l l p{0.27\textwidth} p{0.40\textwidth}@{}}
\toprule
Phase & Query time & Nominal time & Corresponding metric \\
\midrule
Approach & Far approach & $t_{\mathrm{ease}}-1.5$\,s & Pre-emergence unnecessary-braking rate \\
Approach & Deceleration onset & $t_{\mathrm{ease}}$ & Anticipatory Deceleration, pre-emergence unnecessary-braking rate \\
Approach & Mid-deceleration & $(t_{\mathrm{ease}}+t_{\mathrm{done}})/2$ & Anticipatory Deceleration, pre-emergence unnecessary-braking rate \\
Critical & Pre-emergence & $t_{\mathrm{rev}}-0.5$\,s & Primary metric, pre-emergence unnecessary-braking rate \\
Critical & Emergence & $t_{\mathrm{rev}}$ & Primary metric, Braking Response \\
Critical & Post-emergence & $t_{\mathrm{rev}}+0.25$\,s & Primary metric \\
Critical & Braking onset & $t_{\mathrm{brake}}$ & Primary metric \\
Critical & Braking 1/3 and 2/3 & $t_{\mathrm{brake}}+\frac{k}{3}(t_{\mathrm{stop}}-t_{\mathrm{brake}})$, $k\in\{1,2\}$ & Primary metric \\
Critical & Braking midpoint & $(t_{\mathrm{brake}}+t_{\mathrm{stop}})/2$ & Primary metric \\
Stopped & Stopped & $t_{\mathrm{stop}}$ & Remaining Stopped \\
Stopped & Waiting & $(t_{\mathrm{stop}}+t_{\mathrm{resume}})/2$ & Remaining Stopped \\
Resumption & Resumption & $t_{\mathrm{resume}}$ & Resuming Motion \\
Paired control & Four control queries & $t_{\mathrm{rev}}$, $t_{\mathrm{rev}}+0.5$\,s, $t_{\mathrm{rev}}+1.0$\,s, $\bar{t}_{\mathrm{resume}}$ & Unnecessary Braking Rate and Control-Scenario Progress. The emergence query is also used to compute Braking Response. \\
\bottomrule
\end{tabular}
\end{table*}

\begin{table*}[t]
\centering
\caption{Model inputs and trajectory outputs used for training and evaluation. The symbol $t$ denotes the current query time.}
\label{tab:model_io}
\footnotesize
\begin{tabular}{@{}p{0.13\textwidth} p{0.25\textwidth} p{0.26\textwidth} p{0.25\textwidth}@{}}
\toprule
Configuration & AutoVLA & ReCogDrive & LAW \\
\midrule
Camera input & Front, front-left, and front-right & Front & Front, front-right, and front-left \\
Image preprocessing & Rendered frames at $800\times450$ & Upsampled to $1920\times1080$ & Resized to $1600\times900$ and subsequently scaled by $0.4$ \\
Temporal input & Four frames per camera at $t-1.5$, $t-1.0$, $t-0.5$, and $t$\,s & Current front-view image and ego poses at $t-1.5$, $t-1.0$, $t-0.5$, and $t$\,s & Images at $t-1.5$ and $t$\,s \\
Ego-motion state & Current speed and acceleration magnitudes included in the text prompt & Current longitudinal and lateral velocity and acceleration components & The trajectory head uses visual input and the navigation command \\
Trajectory output & 10 waypoints over $5$\,s, with position and heading encoded as action tokens & 8 waypoints over $4$\,s, each containing position and heading & 6 incremental planar displacements over $3$\,s \\
Output coordinate system & Current ego frame & Current ego frame & Current LiDAR frame \\
\bottomrule
\end{tabular}
\end{table*}

\subsection{Capability Metrics and Constraints for Occluded-Pedestrian Emergence}
\label{sec:appendix_metrics}

\noindent\textbf{Query times and evaluation phases.}
Open-loop evaluation is conducted at the nominal query times defined in Appendix~B.5.
Let $t_{\mathrm{ease}}$ and $t_{\mathrm{done}}$ denote the times at which the expert begins anticipatory deceleration and reaches $v_{\mathrm{safe}}$, respectively.
The pedestrian emergence time is denoted by $t_{\mathrm{rev}}$, and the expert begins braking at $t_{\mathrm{brake}} = t_{\mathrm{rev}} + \tau_r$.
In the target scenario, $t_{\mathrm{stop}}$ and $t_{\mathrm{resume}}$ denote the times at which the expert comes to a stop and resumes motion.
The corresponding resumption time for the paired control is denoted by $\bar{t}_{\mathrm{resume}}$.
The same scoring protocol is applied to the development and sealed test sets.

\noindent\textbf{Open-loop collision checking and primary metric.}
Starting from the current ego pose, we linearly interpolate each segment of the predicted waypoint sequence and divide every $0.5$\,s waypoint interval into five $0.1$\,s substeps.
The ego vehicle is represented by a $4.7$\,m $\times$ $1.9$\,m oriented rectangle whose heading follows the corresponding trajectory segment.
The pedestrian is represented by a circular footprint with a radius of $0.3$\,m and moves along the prespecified path from its scripted departure time.
A query is classified as a collision if the two footprints intersect or touch at any substep within the prediction horizon.
The primary metric $F_1$ is the Critical-Phase Collision-Free Rate, defined as one minus the fraction of critical-phase queries classified as collisions.
The critical phase begins $0.5$\,s before pedestrian emergence and comprises the queries labeled Critical in Table~\ref{tab:query_times}.
The sealed test set contains 1,089 critical-phase queries.

\noindent\textbf{Behavior-specific metrics.}
Predicted and expert trajectories are evaluated over the model-specific prediction horizon.
Let $v^{(1)}_t$ denote the mean speed over the first $0.5$\,s segment of the predicted trajectory, $d_t(h)$ denote its displacement over the first $h$ seconds, and $\ell_t$ denote its arc length over the complete prediction horizon.
\begin{itemize}
\item \textbf{Anticipatory Deceleration (AD)} is the fraction of predictions at the deceleration-onset and mid-deceleration queries in which the mean speed of the $0.5$\,s segment containing $t_{\mathrm{done}}$ is at most $4.5$\,m/s.
If $t_{\mathrm{done}}$ lies beyond the prediction horizon, the final segment is used.
\item \textbf{Braking Response (BR)} is the fraction of pairs $(s, \bar{s})$ that satisfy $\ell_t(\bar{s}) - \ell_t(s) > 1$\,m at the emergence query.
\item \textbf{Remaining Stopped (RS)} is the fraction of predictions at the stopped and waiting queries that satisfy $d_t(1) < 0.5$\,m.
\item \textbf{Resuming Motion (RM)} is the fraction of predictions at the resumption query that satisfy $\ell_t > 3$\,m.
\item \textbf{Unnecessary Braking Rate (UBR)} is the fraction of predictions at the four paired-control queries that satisfy $v^{(1)}_t < \bar{v}^{\star(1)}_t - 1$\,m/s, where $\bar{v}^{\star(1)}_t$ is the expert's mean speed over the first predicted segment in the paired control.
\item \textbf{Control-Scenario Progress (CSP)} is the mean of $\min\{\ell_t / \bar{\ell}^{\star}_t,\, 1\}$ over the same four queries, where $\bar{\ell}^{\star}_t$ is the arc length of the paired-control expert trajectory over the same prediction horizon.
\end{itemize}

\noindent\textbf{Checkpoint eligibility constraints.}
An open-loop checkpoint is eligible only if its development-set Unnecessary Braking Rate is at most $0.10$ and its Remaining Stopped score is at least $0.70$.
AutoVLA and ReCogDrive must additionally achieve a pre-emergence unnecessary-braking rate of at most $0.10$.
This rate uses the same first-segment speed criterion, with the target-scenario expert trajectory as the reference, and is evaluated at far approach, deceleration onset, mid-deceleration, and pre-emergence.
LAW uses camera inputs without ego state (Table~\ref{tab:model_io}), so the speed of its first predicted segment fails to track the current ego speed, and this constraint is therefore omitted for LAW.
The original-task validation error $L_{\mathrm{val}}$ is computed on real observations from the nuScenes validation split and must satisfy $L_{\mathrm{val}}(\pi) \le 1.10\, L_{\mathrm{val}}(\pi_0)$, where $\pi_0$ is the corresponding base checkpoint.
AutoVLA uses the cumulative mean L2 error from $1$ to $3$\,s, ReCogDrive uses its official L1 error, and LAW uses the mean L2 error at $1$, $2$, and $3$\,s.

\noindent\textbf{Failure map.}
A target event is classified as a failure if the predicted trajectory collides with the pedestrian at any critical-phase query.
In Eq.~(3), $m_r(b)$ denotes the fraction of failed events in bin $b$ of the development set.

\noindent\textbf{Closed-loop metrics.}
Closed-loop evaluation is conducted at the episode level using the simulation protocol described in Appendix~B.7.
Pedestrian Collision Rate is the fraction of target episodes terminated by a pedestrian collision.
Safe-Passage Rate is the fraction of target episodes in which the rear of the ego vehicle passes the conflict point without any terminal event.
No-Event Rate is the fraction of paired-control episodes completed without any terminal event.
Episode progress is defined as the cumulative forward arc length traveled by the ego vehicle within the scoring window $[0, T_e)$, divided by the corresponding arc length of the rule-based expert.
The ratio is clipped at $1$.
Mean progress is reported separately for the target and paired-control episodes.
The closed-loop primary score is one minus the Pedestrian Collision Rate.
A closed-loop checkpoint must satisfy the same $L_{\mathrm{val}}$ constraint.
Its paired-control progress must also be at least 90\% of the corresponding measurement obtained by the checkpoint used to initialize closed-loop post-training under the same evaluation setup.

\subsection{Closed-Loop Post-Training for Occluded-Pedestrian Emergence}
\label{sec:appendix_closed_loop}

This section details the closed-loop post-training procedure introduced in Sec.~4.1.
The three models share the episode pools, the simulator, the rule-based expert, and the training procedure.

\noindent\textbf{Episode pools.}
The closed-loop training, development, and sealed-test pools are built from the events generated for the open-loop experiments and follow the same log-level scene partition.
Events are excluded if the rule-based expert fails to clear the conflict point within the $15$\,s episode limit or if the required observation history is unavailable.
The three pools contain 762, 348, and 326 episodes, respectively, corresponding to 381, 174, and 163 target--control pairs drawn from 88, 20, and 19 base scenes.
The development and sealed-test pools remain fixed throughout closed-loop post-training.

\noindent\textbf{Policy execution and pedestrian triggering.}
At each $0.5$\,s simulation step, the renderer generates the camera observations of the model from the current ego pose, the policy predicts its future waypoints, and the simulator executes only the first waypoint before replanning.
The model inputs take the same form as in the open-loop evaluation (Appendix~B.5).
The pedestrian is triggered when the ego vehicle reaches the position of the recorded ego vehicle at the nominal pedestrian-departure time, rather than at a fixed simulation time.

\noindent\textbf{Episode horizon and terminal events.}
Let $T_e$ denote the time $2$\,s after the rule-based expert's rear bumper clears the conflict point.
A rollout continues until the later of $T_e$ and $2$\,s after the policy-controlled ego vehicle clears the conflict point, with a limit of $15$\,s.
An episode terminates upon a pedestrian collision, an occluding-vehicle collision, a lateral deviation of more than $2$\,m from the route centerline, a heading deviation of more than $40^\circ$ from the route tangent, or a longitudinal velocity below $-0.5$\,m/s.

\noindent\textbf{Training procedure.}
Closed-loop post-training follows a DAgger-style data-aggregation procedure~\cite{ross2011reduction}, in which the current policy is rolled out in the simulator on the training pool and the rule-based expert provides its plan from each policy-visited state as a corrective target.
Training uses only the states in the task-relevant region, which extends from $20$\,m before the recorded emergence position to the conflict point.
Each such state is paired with four real samples from the nuScenes training split, so the original-task replay proportion remains $\rho = 0.8$.
Training remains supervised and uses no reward.
The policy-visited states are supervised by the expert plans, and the real samples are supervised by the recorded trajectories.
Each model updates the same parameters as in open-loop post-training (Appendix~B.4) and keeps the same learning rate, except that the learning rate of ReCogDrive is reduced to $1\times10^{-6}$.

\noindent\textbf{Checkpoint evaluation.}
The checkpoints of all three models are evaluated on the fixed development pool, with one minus the pedestrian collision rate as the primary metric and the two closed-loop constraints of Appendix~B.6.

\subsection{Instantiation of the Stop-Sign-Controlled Intersection Capability}
\label{sec:appendix_stop_sign}

This section describes how we instantiate the data flywheel for stopping and yielding at stop-sign-controlled intersections (Sec.~4.3).
Unless otherwise specified, the experimental settings follow those of the occluded-pedestrian capability.

\noindent\textbf{Scenario template.}
The base scenes are unsignalized intersections reconstructed from nuScenes logs.
We consider two-way stop control, where stop signs govern only one road and cross traffic has the right of way.
In Type-A scenes, the ego vehicle travels on the stop-controlled road and must stop before the stop line, yield to cross traffic, and resume motion after the conflict has cleared.
In Type-B scenes, the stop sign and stop line are placed on the cross street, and the ego vehicle should proceed without unnecessary braking.
Type-B scenes serve as control scenarios for this capability.
Type-A and Type-B scenes are constructed from different base intersections.
A1 and B1 contain no cross traffic.
A2 inserts scripted conflict vehicles with the right of way on the cross street, and the ego vehicle must continue yielding until they clear the conflict point.
B2 inserts a vehicle on the stop-controlled cross street that stops and yields to the ego vehicle.

\noindent\textbf{Stop signs and stop lines.}
We retain existing stop signs when a usable sign is present in the original scene.
Otherwise, we insert a textured stop-sign asset at the roadside position that is most visible to the approaching ego vehicle.
The stop line is the line before which the ego vehicle must stop.
We use a visible stop line when one is present.
If a crosswalk is visible but no stop line is present, we use the edge of the crosswalk closest to the ego vehicle as the stop line.
If neither is available, we insert a visible stop line.

\noindent\textbf{Scenario parameters.}
The scenario parameters of an A2 event include the number and approach directions of the conflict vehicles, vehicle assets, vehicle speeds, and the arrival offset of the first conflict vehicle relative to the time at which the ego vehicle comes to a stop.
Conflict-vehicle speeds are sampled from $\{6, 9, 12\}$\,m/s.
The arrival offset $\Delta t$ is sampled from $\{0.5, 1.5, 2.5, 3.5\}$\,s.
The four values of $\Delta t$ define the four A2 difficulty bins.
A1, B1, and B2 form three additional condition bins.
The A1 events in the development and sealed test sets include four approach variants formed by combining cruising speeds of $\{5, 8\}$\,m/s with approach decelerations of $\{1.8, 3.0\}$\,m/s$^2$.

\noindent\textbf{Rule-based expert.}
For Type-A scenes, the expert targets a full stop with the front bumper $0.5$\,m before the stop line.
After entering the region within $30$\,m of the stop line, the expert limits its speed according to the remaining stopping distance and applies a braking deceleration of $2.5$\,m/s$^2$.
In A1 scenes, the expert remains stopped for $2$\,s before resuming motion.
In A2 scenes, the expert remains stopped until the rear of the final conflict vehicle has traveled $2$\,m beyond the conflict point, followed by an additional waiting period of $0.8$\,s.
The expert then accelerates toward the recorded speed at $1.5$\,m/s$^2$.
At intersections where the recorded driver already made a complete stop for a real stop sign, we use the recorded trajectory as the expert trajectory.
For Type-B scenes, the expert follows the recorded ego trajectory without braking for the stop sign on the cross street.

\noindent\textbf{Validity checks.}
A Type-A event is retained only if the expert trajectory contains exactly one valid stopping phase.
The stop sign must be reliably detected by at least one forward-facing camera at multiple approach query states.
Each inserted conflict vehicle must reach an image height of at least 25 pixels in at least five frames before arriving at the conflict point.

\noindent\textbf{Data splits.}
All intersections are assigned to splits according to their source logs before training.
The 43 training intersections are drawn from the nuScenes training split, while the development and sealed-test intersections are drawn from the validation split.
The development set contains eight intersections, 230 events, and 1,642 query states.
The sealed test set contains two Type-A intersections and five Type-B intersections, with 248 events and 1,641 query states.
Type-A events constructed from the same intersection and approach trajectory are counted as one independent approach.
The development set contains 20 independent approaches, and the sealed test set contains 10.

\noindent\textbf{Query states.}
For each Type-A event, we sample eight query states along the expert trajectory.
These states occur $5$, $3$, $1.5$, and $0.5$\,s before the expert stop, $0.3$\,s after the stop, at the midpoint of the stopped phase, when the expert resumes motion, and $1$\,s after resumption.
For each Type-B event, we sample five query states at $5$, $3$, $1.5$, and $0.3$\,s before the stop line and $1$\,s after passing it.

\noindent\textbf{Capability metrics.}
All metrics are computed from the predicted trajectory at each query state.
Predicted waypoints at $0.5$\,s intervals are interpolated to $0.1$\,s resolution for evaluating stopping, yielding, and collision behavior.
The primary metric $F_1$ is the Stop-before-Line Rate (SLR), which is computed at the query state $1.5$\,s before the expert stop and averaged over independent approaches.
A prediction is counted as a successful stop if it contains two consecutive $0.5$\,s steps with displacements below $0.05$\,m, the front bumper crosses the stop line by no more than $0.3$\,m before stopping, and the predicted stopping position lies within $5$\,m before the expert stopping position.
The Early Stopping Rate (ESR) measures the fraction of braking predictions whose stopping positions lie more than $5$\,m before the expert stopping position.
The Yielding Rate (YR) is evaluated at A2 query states during the expert's stopped phase.
A prediction is counted as yielding if the ego vehicle advances by no more than $1$\,m before the final conflict vehicle leaves the conflict region.
Resuming Motion (RM) measures whether the predicted trajectory travels at least $3$\,m within the prediction horizon at and after the expert resumption state.
The Conflict-Vehicle Collision Rate (CVCR) measures whether the predicted ego trajectory intersects a scripted conflict vehicle within the prediction horizon.
The Unnecessary Braking Rate (UBR) is computed in the Type-B scenes.
At query states where the current speed is at least $2$\,m/s and the expert does not decelerate, a prediction is counted as unnecessary braking if its $3$\,s travel distance is below 70\% of the distance obtained by maintaining the current speed, or if its average speed over any $1$\,s window falls below 60\% of the current speed.

\noindent\textbf{Checkpoint constraints.}
The Unnecessary Braking Rate of a candidate checkpoint must not exceed the corresponding base-model value by more than $0.05$.
The resuming-motion rate must be at least $0.90$, the early-stopping rate must not exceed $0.10$, and the original-task validation error $L_{\mathrm{val}}$ must remain within $1.10$ times the corresponding base-model value.
For experiments initialized from the occluded-pedestrian checkpoint, we additionally evaluate the previously acquired capability and report its critical-phase collision-free rate as CFR$_{\mathrm{prev}}$.

\noindent\textbf{Event generation and post-training.}
We generate 250 events in the first round and 200 new events in each subsequent round.
In each round, about two thirds of the generation budget is assigned to Type-A events and the remainder to Type-B events.
Within each type, Eq.~(3) allocates the budget across A1, the four A2 difficulty bins, B1, and B2.
A Type-A event is marked as failed if the model fails to stop at any braking query state, fails to yield at any stopped-phase query state, or predicts a collision with a conflict vehicle.
A Type-B event is marked as failed if unnecessary braking occurs at any query state.
AutoVLA and ReCogDrive use the same supervised objectives and optimization settings as in Appendix~B.4.
Samples of the occluded-pedestrian capability account for 50\% of the generated-data pool, and the original-task replay proportion is $\rho = 0.5$.

\noindent\textbf{Checkpoint selection.}
After each epoch, we evaluate the checkpoint on the fixed development set and compute the original-task validation error $L_{\mathrm{val}}$.
Among the checkpoints satisfying the constraints above, we select the checkpoint with the highest primary metric on the development set as the candidate for the current round.
A candidate replaces the retained model only if it improves the development-set stop-before-line rate by at least one percentage point.
Otherwise, the retained model is carried forward.
The sealed test set is used only for final reporting.

\providecommand{\CVal}[2]{#1\,{\footnotesize ($#2\%$)}}

\section{Additional Ablation Studies}
\label{sec:appendix_ablations}

\noindent\textbf{Failure Feedback and Multi-Round Post-Training.}
Table~\ref{tab:budget_numbers} provides the numerical results shown in Figure~3 of the main paper, and Table~\ref{tab:budget_full} reports the full set of metrics at the 100k-step budget.
All six schemes satisfy the selection constraints, and the metrics other than the primary metric show no consistent ordering across the schemes.

\begin{table}[t]
\centering
\caption{Ablation of failure feedback and multi-round post-training on ReCogDrive.
We report critical-phase collision-free rates on the sealed test set, with the best result at each training budget in bold.
FF, UQ, and Rand denote quotas based on a frozen failure map, uniform quotas, and random sampling, and MR and SS denote multi-round and single-stage post-training.}
\label{tab:budget_numbers}
\small
\setlength{\tabcolsep}{5pt}
\renewcommand{\arraystretch}{0.92}
\begin{adjustbox}{max width=\columnwidth,center}
\begin{tabular}{@{}lccccc@{}}
\toprule
Scheme & 20k & 40k & 60k & 80k & 100k \\
\midrule
Rand, SS & 0.614 & 0.640 & 0.676 & 0.676 & 0.676 \\
Rand, MR & 0.624 & 0.667 & 0.667 & 0.691 & 0.691 \\
UQ, MR & 0.624 & 0.667 & 0.691 & 0.691 & 0.691 \\
UQ, SS & 0.594 & 0.627 & 0.667 & 0.675 & 0.695 \\
FF, MR & 0.624 & 0.667 & \textbf{0.692} & \textbf{0.723} & 0.723 \\
Ours & 0.624 & 0.667 & \textbf{0.692} & 0.721 & \textbf{0.751} \\
\bottomrule
\end{tabular}
\end{adjustbox}
\end{table}

\begin{table}[t]
\centering
\caption{Full set of metrics at the 100k-step budget for the six schemes in Table~\ref{tab:budget_numbers}.
Parentheses report the relative change in $L_{\mathrm{val}}$ from the base model.}
\label{tab:budget_full}
\small
\setlength{\tabcolsep}{2.8pt}
\renewcommand{\arraystretch}{0.92}
\begin{adjustbox}{max width=\columnwidth,center}
\begin{tabular}{@{}lcccccccc@{}}
\toprule
Scheme
& \textbf{CFR} $\uparrow$
& AD $\uparrow$
& BR $\uparrow$
& RS $\uparrow$
& RM $\uparrow$
& UBR $\downarrow$
& CSP $\uparrow$
& $L_{\mathrm{val}}\downarrow$ \\
\midrule
Rand, SS & 0.676 & 0.592 & 0.175 & 0.989 & 1.000 & 0.001 & 0.764 & \CVal{0.376}{+2.7} \\
Rand, MR & 0.691 & 0.392 & 0.145 & 0.979 & 1.000 & 0.001 & 0.702 & \CVal{0.389}{+6.3} \\
UQ, MR & 0.691 & 0.514 & 0.165 & 0.997 & 1.000 & 0.004 & 0.694 & \CVal{0.377}{+3.1} \\
UQ, SS & 0.695 & 0.528 & 0.150 & 0.989 & 1.000 & 0.001 & 0.710 & \CVal{0.378}{+3.2} \\
FF, MR & 0.723 & 0.653 & 0.130 & 0.995 & 1.000 & 0.001 & 0.665 & \CVal{0.381}{+4.0} \\
Ours & 0.751 & 0.567 & 0.165 & 0.989 & 1.000 & 0.003 & 0.669 & \CVal{0.386}{+5.4} \\
\bottomrule
\end{tabular}
\end{adjustbox}
\end{table}

\noindent\textbf{Original-Task Data Replay.}
We run four rounds of post-training on ReCogDrive while varying the replay proportion $\rho$ of the original-task data.
Results are reported in Table~\ref{tab:replay_ablation}.
At $\rho=0$, the primary metric score increases from 0.390 for the base model to 0.796, while $L_{\mathrm{val}}$ increases by 41.3\%, indicating that the target capability improvement is accompanied by substantial degradation on the original tasks.
With 50\% replay, the primary metric score remains high at 0.767, while the increase in $L_{\mathrm{val}}$ falls to 4.3\%, within the prescribed 10\% retention tolerance.
Increasing the replay proportion to 67\% and 80\% yields primary metric scores of 0.744 and 0.751, together with $L_{\mathrm{val}}$ increases of 4.6\% and 5.4\%, respectively.
These results show that original-task data replay effectively limits forgetting.

\begin{table}[t]
\centering
\caption{Ablation of original-task data replay on ReCogDrive.}
\label{tab:replay_ablation}
\small
\setlength{\tabcolsep}{2.8pt}
\renewcommand{\arraystretch}{0.92}
\begin{adjustbox}{max width=\columnwidth,center}
\begin{tabular}{@{}lcccccccc@{}}
\toprule
Replay $\rho$
& \textbf{CFR} $\uparrow$
& AD $\uparrow$
& BR $\uparrow$
& RS $\uparrow$
& RM $\uparrow$
& UBR $\downarrow$
& CSP $\uparrow$
& $L_{\mathrm{val}}\downarrow$ \\
\midrule
Base & 0.390 & 0.297 & 0.045 & 0.731 & 0.995 & 0.001 & 0.884 & 0.366 \\
0\% & 0.796 & 0.572 & 0.130 & 0.995 & 1.000 & 0.001 & 0.670 & \CVal{0.517}{+41.3} \\
50\% & 0.767 & 0.450 & 0.140 & 0.981 & 1.000 & 0.001 & 0.635 & \CVal{0.382}{+4.3} \\
67\% & 0.744 & 0.589 & 0.140 & 0.973 & 1.000 & 0.001 & 0.678 & \CVal{0.383}{+4.6} \\
80\% & 0.751 & 0.567 & 0.165 & 0.989 & 1.000 & 0.003 & 0.669 & \CVal{0.386}{+5.4} \\
\bottomrule
\end{tabular}
\end{adjustbox}
\end{table}

\noindent\textbf{Model Acceptance Rule.}
We compare accepting a candidate checkpoint only upon improvement in the primary metric with unconditional acceptance (Table~\ref{tab:acceptance_ablation}).
Under unconditional acceptance, the fourth-round candidate reduces the primary metric on the sealed test set from 0.751 to 0.701.
Anticipatory deceleration and remaining stopped also decrease from 0.567 and 0.989 to 0.489 and 0.947, respectively.
These results show that the acceptance rule prevents regressive updates from propagating to subsequent iterations.

\begin{table}[t]
\centering
\caption{Ablation of the model acceptance rule on ReCogDrive.}
\label{tab:acceptance_ablation}
\small
\setlength{\tabcolsep}{2.8pt}
\renewcommand{\arraystretch}{0.92}
\begin{adjustbox}{max width=\columnwidth,center}
\begin{tabular}{@{}lcccccccc@{}}
\toprule
Acceptance Rule
& \textbf{CFR} $\uparrow$
& AD $\uparrow$
& BR $\uparrow$
& RS $\uparrow$
& RM $\uparrow$
& UBR $\downarrow$
& CSP $\uparrow$
& $L_{\mathrm{val}}\downarrow$ \\
\midrule
Unconditional acceptance & 0.701 & 0.489 & 0.165 & 0.947 & 1.000 & 0.001 & 0.750 & \CVal{0.382}{+4.3} \\
Ours & 0.751 & 0.567 & 0.165 & 0.989 & 1.000 & 0.003 & 0.669 & \CVal{0.386}{+5.4} \\
\bottomrule
\end{tabular}
\end{adjustbox}
\end{table}

\noindent\textbf{Validity Checks.}
The validity checks are applied to the rendered frames of each generated event before it enters the generated-data pool.
An event is excluded if the pedestrian is visible in more than one frame before the specified emergence time, if the emergence time deviates from the specification by more than $0.6$\,s, if the pedestrian is less than 25 pixels tall at emergence, or if an off-the-shelf detector fails to recognize the pedestrian as a person.
Removing the validity checks increases the primary metric from 0.751 to 0.804, while reducing braking response from 0.165 to 0.095 and control-scenario progress from 0.669 to 0.615 (Table~\ref{tab:validity_ablation}).
This indicates that the higher collision-free rate primarily results from more conservative early deceleration rather than a more accurate post-emergence response.
By ensuring that scenarios are presented under the intended conditions, the validity checks reduce the model's reliance on such conservative behavior.

\begin{table}[t]
\centering
\caption{Ablation of validity checks on ReCogDrive.}
\label{tab:validity_ablation}
\small
\setlength{\tabcolsep}{2.8pt}
\renewcommand{\arraystretch}{0.92}
\begin{adjustbox}{max width=\columnwidth,center}
\begin{tabular}{@{}lcccccccc@{}}
\toprule
Validity Checks
& \textbf{CFR} $\uparrow$
& AD $\uparrow$
& BR $\uparrow$
& RS $\uparrow$
& RM $\uparrow$
& UBR $\downarrow$
& CSP $\uparrow$
& $L_{\mathrm{val}}\downarrow$ \\
\midrule
Without checks & 0.804 & 0.583 & 0.095 & 0.973 & 1.000 & 0.001 & 0.615 & \CVal{0.382}{+4.3} \\
Ours & 0.751 & 0.567 & 0.165 & 0.989 & 1.000 & 0.003 & 0.669 & \CVal{0.386}{+5.4} \\
\bottomrule
\end{tabular}
\end{adjustbox}
\end{table}

\section{Details of the Real-Vehicle Evaluation}
\label{sec:appendix_real_vehicle}

\begin{figure*}[t]
\centering
\includegraphics[width=2\columnwidth]{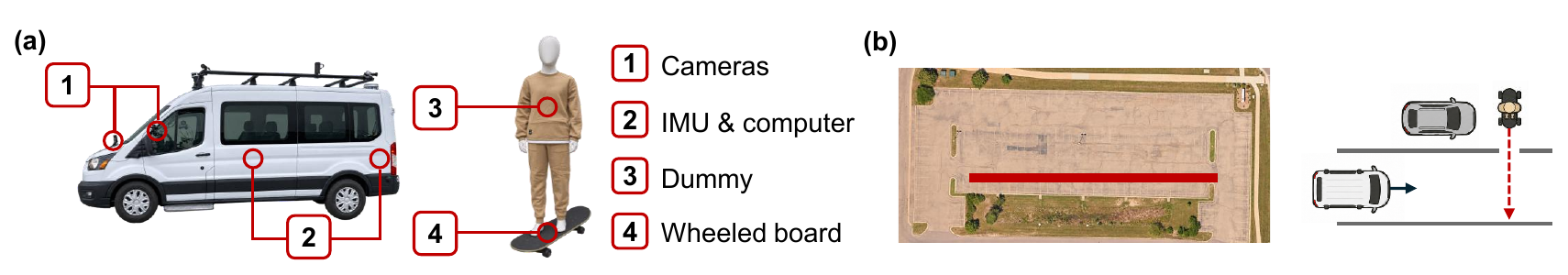}
\caption{Real-vehicle test scenario.
(a) The test vehicle and the pedestrian dummy on a wheeled board.
(b) The test site and the scenario layout, in which the dummy crosses the ego path from behind a parked vehicle.}
\label{fig:real_setup}
\end{figure*}

\noindent\textbf{Experimental setup.}
The test vehicle is an electric connected and autonomous van, a Ford E-Transit, equipped with three forward-facing cameras ($1920\times1080$, $10$\,Hz).
The closed test track is an empty parking lot, where a parked vehicle serves as the occluding vehicle and a child-sized pedestrian dummy on a wheeled board is pulled by a rope from behind it across the ego path (Figure~\ref{fig:real_setup}).
The paired control scenario keeps the occluding vehicle and omits the pedestrian dummy.
ReCogDrive takes the front camera as input and has an inference latency of about $0.8$\,s on the vehicle.
Pedestrian emergence is defined as the first input frame in which the pedestrian dummy is visible.

\begin{figure*}[t]
\centering
\includegraphics[width=\textwidth]{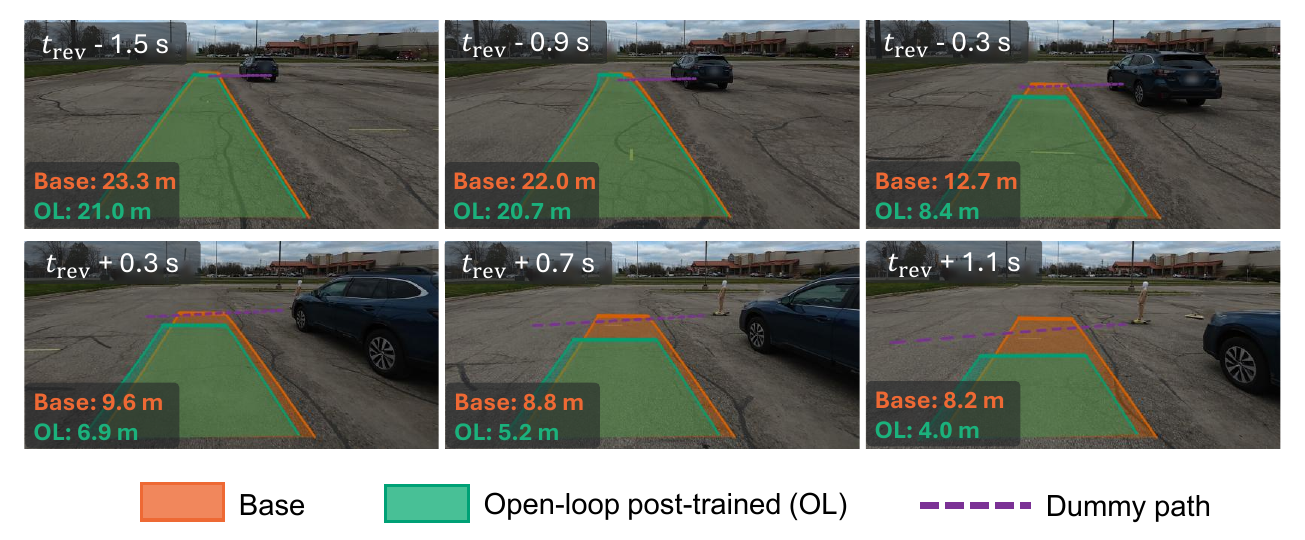}
\caption{ReCogDrive plans on one recorded run.
The top row shows the front-camera frames $1.5$, $0.9$, and $0.3$\,s before pedestrian emergence, when the pedestrian dummy is still occluded, and the bottom row shows the frames $0.3$, $0.7$, and $1.1$\,s after emergence.
The orange and green regions mark the ground area that the front of the vehicle would cover under the $4$\,s plans of the base model and the open-loop post-trained model, respectively, and the purple dashed line marks the subsequent path of the pedestrian dummy.
From $0.3$\,s before emergence onward, the plan of the open-loop post-trained model is clearly shorter than that of the base model and ends before the path of the dummy, whereas the plan of the base model extends onto it.}
\label{fig:real_plans}
\end{figure*}

\noindent\textbf{Open-loop evaluation on the recorded observations.}
The rule-based expert drives the vehicle while the model stays disengaged, so that all checkpoints are queried on identical observations from six target runs and three paired control runs.
The metrics follow Appendix~B.6, where the collision check uses the pedestrian trajectory recovered from the images and the dimensions of the test vehicle, and braking response uses all critical-phase queries after emergence. Figure~\ref{fig:real_plans} shows the plans of the ReCogDrive base and open-loop post-trained models on one of the recorded runs.

Table~\ref{tab:real_open_loop} reports the results of AutoVLA and LAW.
For AutoVLA, open-loop post-training increases the collision-free rate from 0.559 to 0.740 and anticipatory deceleration from 0.220 to 0.576, while the unnecessary braking rate in the paired control rises to 0.128 and control-scenario progress decreases to 0.629.
Its closed-loop checkpoint attains a collision-free rate of 0.575.
The LAW base model barely moves on the recorded observations, with control-scenario progress of 0.023, so its collision-free rate of 1.000 is uninformative.
After post-training, its control-scenario progress rises to 0.249 for the open-loop checkpoint and 0.321 for the closed-loop checkpoint, and its braking response rises from 0.000 to 0.640 and 0.439, while its unnecessary braking rate remains 1.000.

\begin{table}[t]
\centering
\caption{Open-loop results of AutoVLA and LAW on the recorded observations.
Base denotes the pretrained model, and OL and CL denote the final checkpoints from open-loop and closed-loop post-training.}
\label{tab:real_open_loop}
\small
\setlength{\tabcolsep}{2.8pt}
\renewcommand{\arraystretch}{0.92}
\begin{adjustbox}{max width=\columnwidth,center}
\begin{tabular}{@{}llccccccc@{}}
\toprule
Model & Stage
& \textbf{CFR} $\uparrow$
& AD $\uparrow$
& BR $\uparrow$
& RS $\uparrow$
& RM $\uparrow$
& UBR $\downarrow$
& CSP $\uparrow$ \\
\midrule
AutoVLA & Base & 0.559 & 0.220 & 0.623 & 1.000 & 1.000 & 0.000 & 0.969 \\
 & OL & 0.740 & 0.576 & 0.298 & 0.705 & 0.875 & 0.128 & 0.629 \\
 & CL & 0.575 & 0.763 & 0.474 & 0.659 & 0.906 & 0.051 & 0.921 \\
\midrule
LAW & Base & 1.000 & 1.000 & 0.000 & 1.000 & 0.000 & 1.000 & 0.023 \\
 & OL & 1.000 & 1.000 & 0.640 & 0.273 & 0.000 & 1.000 & 0.249 \\
 & CL & 1.000 & 1.000 & 0.439 & 0.182 & 0.438 & 1.000 & 0.321 \\
\bottomrule
\end{tabular}
\end{adjustbox}
\end{table}

\noindent\textbf{Closed-loop checkpoint.}
Table~\ref{tab:cl_sim_open_loop} reports the open-loop metrics of the three ReCogDrive checkpoints on the sealed test set in simulation.
In simulation, the closed-loop checkpoint attains an open-loop collision-free rate of 0.534, compared with 0.765 for the open-loop checkpoint, and its remaining-stopped score decreases from 0.976 to 0.689.
This agrees with its results on the recorded observations.
A possible reason is that the closed-loop simulation executes only the first predicted waypoint.
Closed-loop post-training therefore emphasizes this waypoint, and the quality of the entire predicted trajectory lags behind, whereas the open-loop metrics assess the entire trajectory.

\begin{table}[t]
\centering
\caption{Open-loop results of the three ReCogDrive checkpoints on the sealed test set in simulation.}
\label{tab:cl_sim_open_loop}
\small
\setlength{\tabcolsep}{2.8pt}
\renewcommand{\arraystretch}{0.92}
\begin{adjustbox}{max width=\columnwidth,center}
\begin{tabular}{@{}lccccccc@{}}
\toprule
Stage
& \textbf{CFR} $\uparrow$
& AD $\uparrow$
& BR $\uparrow$
& RS $\uparrow$
& RM $\uparrow$
& UBR $\downarrow$
& CSP $\uparrow$ \\
\midrule
Base & 0.390 & 0.297 & 0.045 & 0.731 & 0.995 & 0.001 & 0.884 \\
OL & 0.765 & 0.631 & 0.185 & 0.976 & 1.000 & 0.001 & 0.733 \\
CL & 0.534 & 0.792 & 0.175 & 0.689 & 1.000 & 0.004 & 0.886 \\
\bottomrule
\end{tabular}
\end{adjustbox}
\end{table}

\section{Bootstrap Confidence Intervals for Table 1}

\begin{table}[t]
\centering
\caption{Bootstrap 95\% confidence intervals of the CFR in Table~1 of the main paper.}
\label{tab:ci_table1}

\definecolor{RFiveBlue}{HTML}{EAF2FB}

\small
\setlength{\tabcolsep}{2.8pt}
\renewcommand{\arraystretch}{0.92}

\begin{adjustbox}{max width=\columnwidth,center}
\begin{tabular}{@{}clcc@{}}
\toprule
Model
& Round
& \textbf{CFR} $\uparrow$
& 95\% CI \\
\midrule

\multirow[c]{6}{*}{\textbf{AutoVLA}}
& Base & 0.261 & [0.214, 0.311] \\
& R1 & 0.396 & [0.352, 0.446] \\
& R2 & 0.519 & [0.470, 0.572] \\
& R3 & 0.544 & [0.501, 0.591] \\
& R4 & 0.578 & [0.534, 0.624] \\
\rowcolor{RFiveBlue}\cellcolor{white}
& R5 & 0.579 & [0.543, 0.617] \\

\midrule

\multirow[c]{6}{*}{\textbf{ReCogDrive}}
& Base & 0.390 & [0.358, 0.426] \\
& R1 & 0.667 & [0.629, 0.705] \\
& R2 & 0.708 & [0.670, 0.745] \\
& R3 & 0.751 & [0.715, 0.787] \\
& R4 & 0.746 & [0.706, 0.785] \\
\rowcolor{RFiveBlue}\cellcolor{white}
& R5 & 0.765 & [0.724, 0.805] \\

\midrule

\multirow[c]{6}{*}{\textbf{LAW}}
& Base & 0.967 & [0.935, 0.993] \\
& R1 & 0.786 & [0.762, 0.808] \\
& R2 & 0.785 & [0.761, 0.807] \\
& R3 & 0.796 & [0.771, 0.818] \\
& R4 & 0.795 & [0.776, 0.812] \\
\rowcolor{RFiveBlue}\cellcolor{white}
& R5 & 0.791 & [0.768, 0.811] \\

\bottomrule
\end{tabular}
\end{adjustbox}
\end{table}
\label{sec:appendix_ci}

Table~\ref{tab:ci_table1} reports the 95\% confidence intervals of the critical-phase collision-free rate (CFR) for each row of Table~1 of the main paper.
The intervals are computed with the percentile bootstrap, in which the 20 base scenes of the sealed test set are resampled with replacement 10,000 times and the CFR is recomputed each time over all critical-phase queries of the resampled scenes.
These intervals reflect the uncertainty due to the sampling of test scenes.
For AutoVLA and ReCogDrive, the interval of every post-training round lies entirely above that of the base model, and the interval of Round~5 lies entirely above that of Round~1.
For LAW, the intervals of the post-training rounds overlap one another and all lie below that of the base model, whose high CFR arises mainly from policy stalling, as discussed in Sec.~4.2 of the main paper.


\end{document}